%% file: main.tex
\documentclass[10pt,twocolumn,letterpaper]{article}

\usepackage[pagenumbers]{iccv} 

\input{preamble}

\definecolor{iccvblue}{rgb}{0.21,0.49,0.74}
\usepackage[pagebackref,breaklinks,colorlinks,allcolors=iccvblue]{hyperref}

\def\paperID{1791} 
\def\confName{ICCV}
\def\confYear{2025}

\title{Progressive Test Time Energy Adaptation for Medical Image Segmentation}

\author{Xiaoran Zhang\textsuperscript{1} \quad
Byung-Woo Hong\textsuperscript{2} \quad
Hyoungseob Park\textsuperscript{1} \quad
Daniel H. Pak\textsuperscript{1} \quad \\
Anne-Marie Rickmann\textsuperscript{1} \quad
Lawrence H. Staib\textsuperscript{1} \quad
James S. Duncan\textsuperscript{1} \quad
Alex Wong\textsuperscript{1}
\vspace{3mm} \\
\textsuperscript{1}Yale University \quad \textsuperscript{2}Chung-Ang University \\
\tt\small \{xiaoran.zhang,james.duncan,alex.wong\}@yale.edu
}

\begin{document}
\maketitle

\input{sec/0_abstract}
\input{sec/1_intro}

\input{sec/2_related_works}
\input{sec/3_methods}

\input{sec/4_experiments}

\input{sec/5_results}
\input{sec/6_conclusion}

{
    \small
    \bibliographystyle{ieeenat_fullname}
    \bibliography{main}
}

\input{sec/X_suppl}

\end{document}

%% file: preamble.tex
\newcommand{\revise}[1]{#1}
\usepackage[ruled,vlined,linesnumbered]{algorithm2e}
\usepackage{adjustbox}
\usepackage{multirow}
\usepackage{pifont}
\newcommand{\cmark}{\textcolor{green}{\ding{51}}}%
\newcommand{\xmark}{\textcolor{red}{\ding{55}}}%


%% file: sec/0_abstract.tex
\begin{abstract}
We propose a model-agnostic, progressive test-time energy adaptation approach for medical image segmentation. Maintaining model performance across diverse medical datasets is challenging, as distribution shifts arise from inconsistent imaging protocols and patient variations. Unlike domain adaptation methods that require multiple passes through target data—impractical in clinical settings—our approach adapts pretrained models progressively as they process test data. Our method leverages a shape energy model trained on source data, which assigns an energy score at the patch level to segmentation maps: low energy represents in-distribution (accurate) shapes, while high energy signals out-of-distribution (erroneous) predictions. 
\revise{By minimizing this energy score at test time, we refine the segmentation model to align with the target distribution. To validate the effectiveness and adaptability, we evaluated our framework on eight public MRI (bSSFP, T1- and T2-weighted) and X-ray datasets spanning cardiac, spinal cord, and lung segmentation. We consistently outperform baselines both quantitatively and qualitatively.} 


\end{abstract}

%% file: sec/1_intro.tex
\section{Introduction}
\label{sec:intro}
Image segmentation aims to partition the image space into semantic categories. While the resulting segmentation maps support a wide range of applications, there is particular interest in medicine as outlining anatomy is not only prohibitively expensive but also requires expert knowledge. \revise{Image segmentation of medical images, e.g., MRI and X-ray 
poses new challenges than those of natural scenes.} Not only are there \revise{fewer} datasets and gold standard labels from experts, but there is also a wide range of nuisances, from heteroscedastic noise and inconsistent imaging protocols to calibration and intensity profiles, causing a \textbf{covariate shift} between data used for training segmentation models and data during testing. This limits the fidelity of medical image analysis to support downstream applications for diagnosis, disease progression monitoring, and surgical guidance.

\begin{figure}[tb]
    \centering
    \includegraphics[scale=0.06]{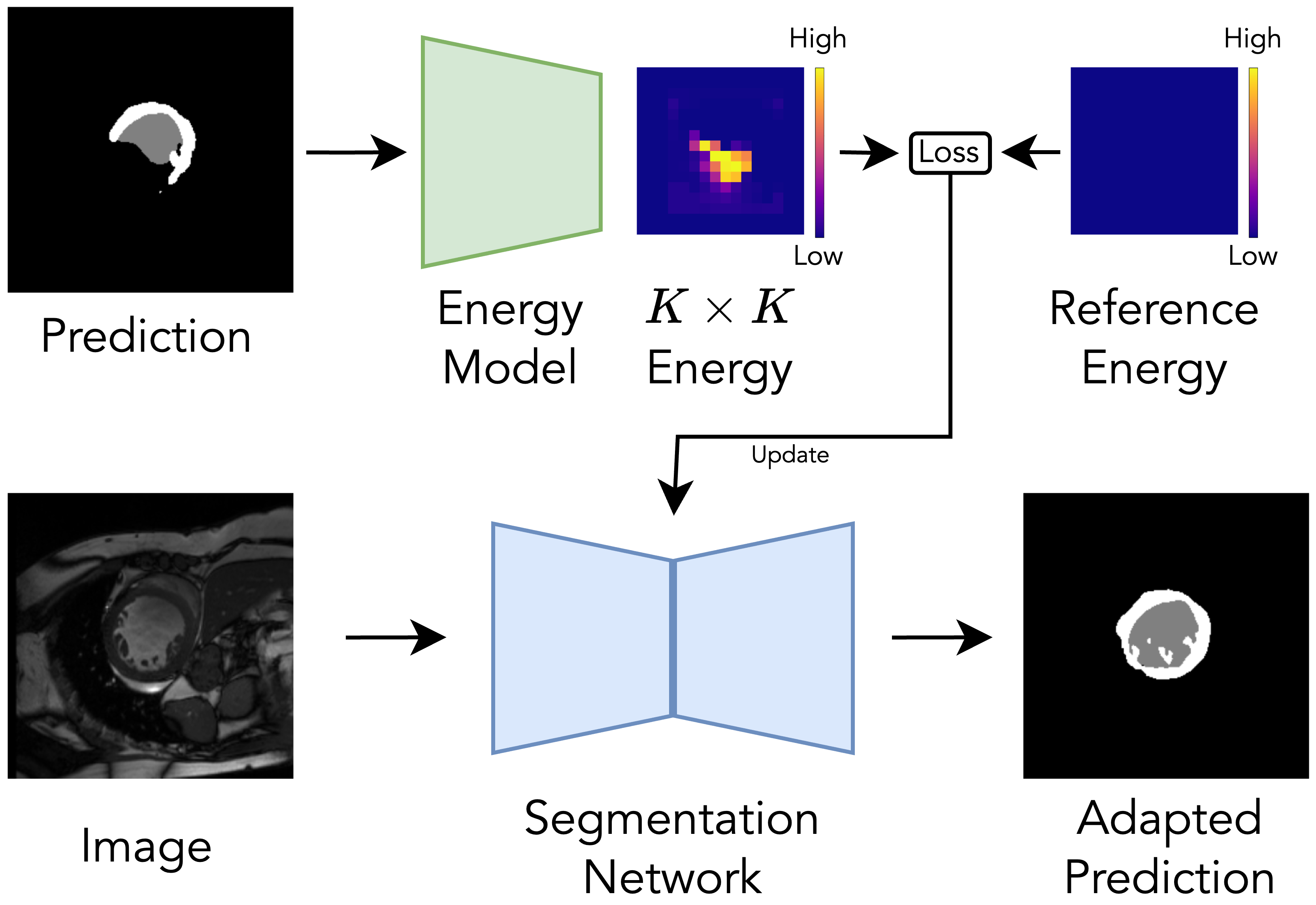}
    \caption{Illustration of a single-step update in our proposed progressive test-time adaptation framework. The energy model identifies erroneous regions in the initial segmentation prediction by assigning high energy values to these areas. Subsequently, we update the segmentation network to minimize the assigned energy, refining the prediction and improving segmentation accuracy.}
    \label{fig:teaser}
\end{figure}

To mitigate errors \revise{from the covariate shift}, one may adapt or update the model trained on a ``source'' data distribution or domain to fit a ``target'' testing distribution. Existing techniques may assume access to the source domain training data paired with ground truth for supervision after \revise{deployment} \cite{singhal2023domain,nelakurthi2018source}. Variants of these approaches may reduce these assumptions, such as removing the need to have access to training data (source-free) or to ground truth (unsupervised) during testing \cite{ganin_unsupervised_2015, peng_moment_2019}. Nonetheless, they still assume access to a pre-collected dataset of the testing dataset, which is impractical as one cannot anticipate patients and requires many passes through the target testing data; this limits their application in real-time \revise{decision-making.} 

More practically, we will exploit the data observed during test time to close the domain gap. Like domain adaptation, we assume access to a model (pre)trained on source data; unlike them, we do not have access to the source data once deployed. Instead, we will focus on test-time adaptation (TTA), the problem of adapting pretrained models as they process incoming test data, e.g., a single image or a small batch of images within short intervals, as a stream without the ability to revisit previously observed examples. Due to the use of comparatively small batch sizes and the lack of labels (data term), the update of these highly parameterized functions necessitates a strong prior.

Inspired by classical energy-based methods, which aim to implicitly model data distributions by defining an energy function over the data, we propose to utilize them as a test-time adaptation mechanism. This is facilitated by associating a scalar (energy) to data, where low energy depicts more probable (within distribution) data points. Hence, should a segmentation model predict erroneous (out-of-distribution) segmentation maps with shapes exhibiting low fidelity to natural anatomy, the segmentation map would yield high energy. Given a trained energy model, frozen at test time, the adaptation mechanism comes from updating the segmentation model to reduce energy. This is unlike the classical energy-based methods that update the input. 

To this end, we propose a progressive energy-based test-time adaptation method for medical image segmentation. \revise{Following a common practice in TTA \cite{park_test-_2024,song2023ecotta}, our method first trains an energy model on source domain data as a preparation stage.} To ensure that updates are localized, we utilize a region-based energy model acting as a discriminator \revise{that takes a segmentation map and returns energy values for each region.} 
However, \revise{since we have only source data}, we lack out-of-distribution examples. One possibility is to train a generative model \cite{yuan_tea_2024}, but it can be expensive. Instead, we utilize adversarial perturbations to probe the pretrained segmentation model to curate out-of-distribution outputs, allowing us to freely explore the input data space of the energy model. \revise{Once trained}, the energy model is frozen and deployed along with the segmentation model to the test-time target domain. 
\revise{For each new example, it progressively updates segmentation model by minimizing energy, refining predictions as shown in \cref{fig:teaser}.} 
\revise{We evaluate our method on eight public datasets covering cardiac, spinal cord, and lung segmentation from bSSFP, T1-, T2-weighted MRI, and X-rays, consistently outperforming baselines.
}


\revise{\textbf{Our contributions}: (1) We propose the first energy-based formulation for TTA in medical image segmentation, leveraging adversarial perturbations to explore the energy landscape and enforce shape priors for adapting to covariate shifts. (2) Our shape energy model exhibits strong out-of-distribution detection capabilities at test time, accurately identifying erroneous segmentation regions over 92\% accuracy. (3) We consistently outperforms baselines across diverse medical datasets, achieving state-of-the-art results. 
}

%% file: sec/2_related_works.tex
\section{Related Works}
\label{sec:related_works}
\textbf{Medical Image Segmentation} has seen significant advancements in a supervised setting, driven by recent developments in deep learning \cite{ronneberger_u-net_2015, bernard_deep_2018, leclerc_deep_2019, zhang_fully_2021, isensee_nnu-net_2021}. Current methods can be broadly categorized into CNN-based \cite{ronneberger_u-net_2015, roy_mednext_2023}, Transformer-based \cite{hatamizadeh_swin_2022, chen_transunet_2024}, and Mamba-based approaches \cite{ma_u-mamba_2024, wang_mamba-unet_2024}. For CNNs, the UNet architecture \cite{ronneberger_u-net_2015} has proven effective across numerous segmentation tasks \cite{zhang_automatic_2021, li_myops_2023}. Expanding on this, MedNext \cite{roy_mednext_2023} incorporates scalable ConNeXt blocks inspired by Transformer \cite{vaswani_attention_2017} into a UNet structure. For Transformer-based methods, SwinUNETR \cite{hatamizadeh_swin_2022} adopts a U-shaped network with the Swin Transformer \cite{liu_swin_2021} as the encoder to capture detailed features across multiple scales. Leveraging the long-range modeling capabilities of Mamba \cite{gu_mamba_2024}, \citet{ma_u-mamba_2024} introduce mamba layers within skip connections to enhance segmentation performance in complex contexts. To automate critical design choices, such as data processing and model architecture, nnUNet \cite{isensee_nnu-net_2021, isensee_nnu-net_2024} offers a self-configuring framework, streamlining the segmentation pipeline. Although the methods above achieve state-of-the-art performance, they often struggle with distribution shifts \cite{guan_domain_2022, weihsbach_dg-tta_2024} that are prevalent in real-world applications, resulting from factors like varying imaging protocols across institutions and patient-specific differences such as variations in demographics.

\textbf{Domain Adaptation} aims to address distribution shifts between training and testing data, enabling a model to generalize effectively from one domain to another. It involves training on labeled source data and then adapting the model to perform well on target data with a different distribution, often using techniques that require little to no labeled data from the target domain. Related fields include unsupervised domain adaptation \cite{ganin_unsupervised_2015, peng_moment_2019, li_model_2020}, which leverages source domain data, and source-free domain adaptation \cite{kim_domain_2021}, which minimizes source-target discrepancy through multiple passes over the test data in the absence of source data. A number of works have explored to use of domain adaptation to close the performance gap in medical segmentation: \citet{perone_unsupervised_2019} introduced self-ensembling in an unsupervised domain adaptation setting, while \citet{yang_source_2022} proposed Fourier style mining to enable source-free domain adaptation.
However, both approaches rely on prior access to either the test distribution \cite{guan_domain_2022} or the complete test set \cite{yang_source_2022}, which is impractical in real-world clinical settings where patient data cannot be anticipated in advance. 


\textbf{Test-Time Adaptation} (TTA) aims to adapt a model, pretrained on source data, to perform effectively on test data from a target distribution without requiring access to the original source training data.  
Unlike test-time training (TTT) \cite{sun_test-time_2019, sun_test-time_2020, liu_ttt_2021, bartler_mt3_2022}, which typically incorporates an auxiliary self-supervised task (i.e., rotation prediction \cite{sun_test-time_2019}) alongside the primary task to align features, we focus on test-time adaptation in scenarios where the test data is processed in a single pass. Previous approaches have explored various strategies for test-time adaptation. BN \cite{schneider_improving_2020} proposes replacing batch normalization activation statistics from the training set with those from the test data. TENT \cite{wang_tent_2020} introduce an entropy-minimization objective to update BatchNorm layer statistics using test samples, aiming to improve adaptation. EATA \cite{niu_efficient_2022} employs an adaptive weighting strategy, constraining select model parameters from undergoing drastic updates by estimating Fisher importance from test samples. SAR \cite{niu_towards_2023} incorporates sharpness-aware regularization to handle large, noisy gradients arising from entropy minimization, enhancing stability and reliability. Beyond entropy-based methods, CoTTA \cite{wang_continual_2022} uses weight-averaged and augmentation-averaged predictions to generate pseudo labels for adaptation. Similarly, MEMO \cite{zhang_memo_2022} employs a consistency-based approach by minimizing the model's marginal distribution across augmentations. 
\revise{The closest work to ours is TEA \cite{yuan_tea_2024}, which applies energy-based adaptation to classification by generating image samples during optimization. In contrast, our approach is not a direct extension of TEA to segmentation. Instead, we emphasize the discriminative aspect of energy-based models, using a shape energy model as a shape classifier. Additionally, unlike TEA, which assigns a single global energy value per image, our method employs a region-based energy model for finer granularity.} Although few approaches \cite{hu_fully_2021, valanarasu_--fly_2022, weihsbach_dg-tta_2024} have explored the novel test-time adaptation setting in medical imaging, they often depend on specific segmentation models \cite{valanarasu_--fly_2022} or incorporate explicit regularization \cite{hu_fully_2021}, which limit their ability to unseen diverse medical data. In contrast, our approach leverages a shape energy model that regularizes the estimated energy levels to facilitate effective adaptation.


\textbf{Energy-based Models} are a class of non-normalized probabilistic models \cite{lecun_tutorial_nodate} widely applied in generative modeling \cite{ho_denoising_2020, nichol_improved_2021} to transform distributions. They offer a flexible parameterization that does not require an explicit neural network for sample generation, allowing them to model a broad spectrum of probability distributions \cite{yuan_tea_2024}. For example, \citet{han_divergence_2019} introduced a divergence triangle to jointly train a generator, energy-based model, and inference model. Additionally, \citet{du_implicit_2020} proposed a scalable approach for MCMC-based energy training in neural networks. Beyond generative uses, energy-based models can also function as classifiers by associating low energy with high-confidence predictions \cite{grathwohl_your_2020}. In this work, we propose a shape energy model that can be interpreted as a region-based classifier to minimize the estimated energy of prediction for test-time segmentation adaptation.


%% file: sec/3_methods.tex
\section{Methods}
\label{sec:methods}

\begin{figure*}[t]
    \centering
    \includegraphics[scale=0.065]{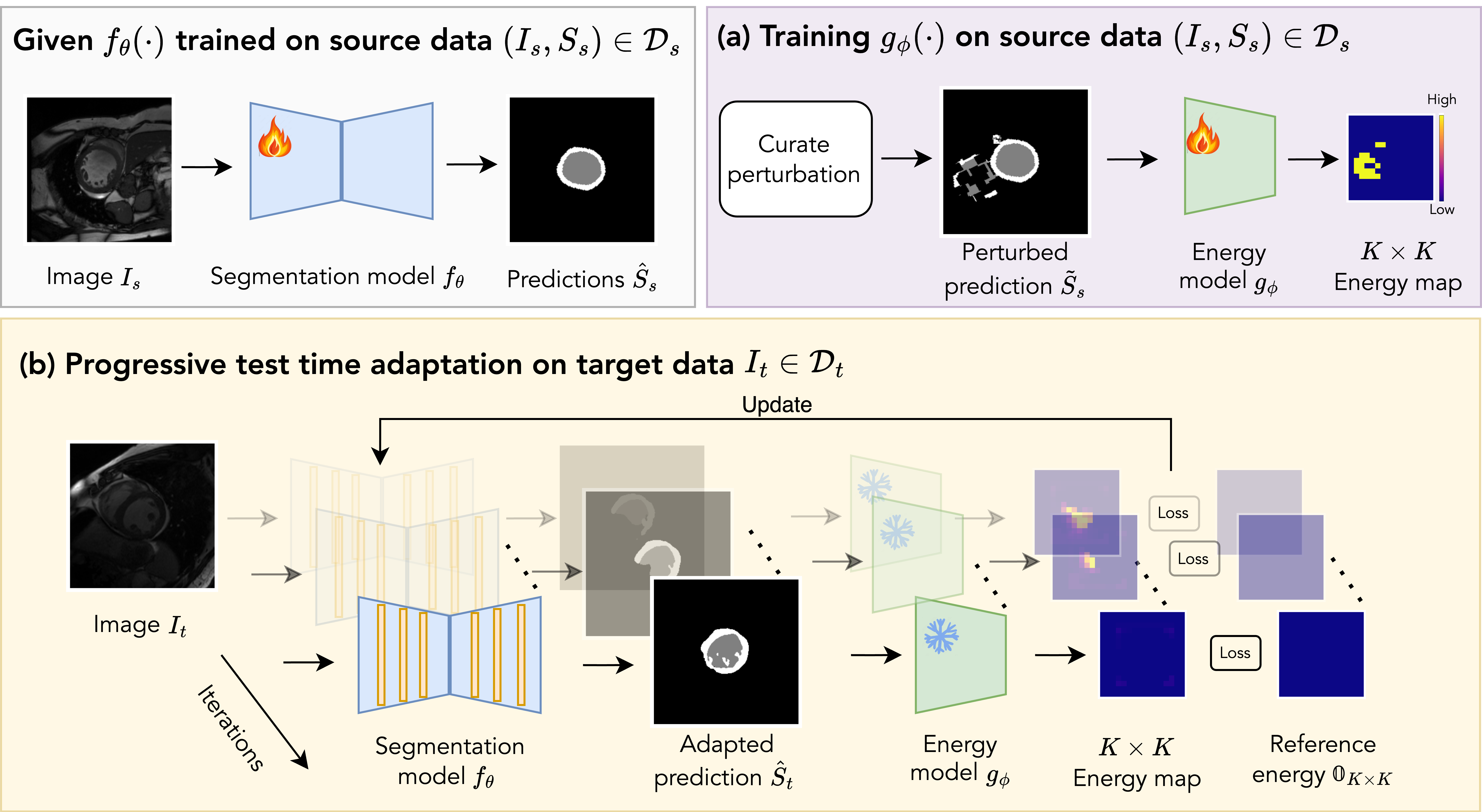}
    \caption{Overview. We assume a segmentation model $f_\theta(\cdot)$ is pretrained on a source dataset. (a) The energy model $g_\phi(\cdot)$ is trained to estimate patchwise energy values, using binary reference energy labels based on the mismatch between perturbed predictions $\hat{Y}_s$ and ground truth shape $Y_s$ on the source dataset. (b) During adaptation, the trained energy model $g_\phi(\cdot)$ is applied to predictions on the test-time distribution, and the BatchNorm layers of $f_\theta(\cdot)$ are updated iteratively to match with uniform low energy as target.}

    \label{fig:framework}
\end{figure*}

\noindent\textbf{Method Formulation.}
Let $I\in\mathbb{R}^{H\times W}$ denote an intensity image, where $I:\Omega\mapsto [0,1]$ is the imaging function and $\Omega$ is the image space. A segmentation network $f_\theta(\cdot)$ aims to output a pixelwise logits $\hat{S} = f_\theta(I) \in \mathbb{R}^{C  \times H \times W}$, where $\theta$ denotes the network parameters and $f: \Omega \mapsto \mathbb{R}^C$ the logits; the per-pixel categories $\bar{S} \in \{0, \dots, C - 1\}^{H \times W}$ are obtained by taking the argmax over the logits, which maps logits to one of $C$ classes to yield a segmentation map. For test-time adaptation, we follow the formulation of \cite{wang_tent_2020, yuan_tea_2024}, assuming that the segmentation network $f_\theta(\cdot)$ is trained solely on the source dataset $\mathcal{D}_s = \{I_s^{(i)}, S_s^{(i)}\}_{i=1}^{N_s}$ without access to the target dataset, where $N_s $ denotes the number of training samples. Our goal is to adapt the network $f_\theta(\cdot)$ to the target dataset $\mathcal{D}_t = \{I_t^{(i)}, S_t^{(i)}\}_{i=1}^{N_t}$, where $N_t$ represents the number of samples in the target dataset. Note: we do not have access to the ground truth $S_t$ for test time adaptation.


To perform adaptation, a test-time objective must be defined to update the source segmentation network $f_\theta(\cdot)$. While general regularization techniques, such as self-entropy minimization, helps to mitigate low-confidence predictions during test time, this approach fails to leverage the strong shape prior inherent in segmentation tasks. However, explicitly defining statistical shape priors and enforcing them is intractable as it requires optimizing shape parameters for each individual object and for all deformations. Yet, one can intuitively determine incorrect predictions (i.e., an incomplete cardiac segmentation in \cref{fig:teaser}) without needing explicit shape parameters by discriminating whether they are within the distribution of plausible shapes. Building on this intuition, we propose to learn a shape energy model, realized as a discriminator or classifier, to distinguish between shapes that are within the distribution of natural anatomy and those that are implausible or out of distribution. 
At test time, the trained shape energy model is used to iteratively update the segmentation model, guiding it to align with the reference energy. \cref{fig:framework} shows an overview of our method. Assuming a segmentation model pretrained on the source dataset, our method comprises two stages: (a) training the shape energy model on the source data, and (b) utilizing the shape energy model to progressively update the segmentation model on the testing data.

\noindent\textbf{Region-based Energy Model.}
Energy models naturally capture distribution changes, making them well-suited for TTA in segmentation. We aim to transform estimated implausible shape distribution to match natural anatomy, e.g., ground truth. It relies on observation \cite{lecun_tutorial_nodate} that any probability distribution for input $x\in\mathbb{R}^D$ can be expressed as 
\begin{align}
    p_\phi(x) = \frac{\exp(-E_\phi(x))}{Z(\phi)},
    \label{eq:energy_formulation}
\end{align}
where $E_\phi(x):\mathbb{R}^D \mapsto \mathbb{R}$ is energy function mapping each point to a scalar and $Z(\phi)=\int_x\exp(-E_\phi(x))$ is normalizing constant. An energy model can be parameterized using any function that takes an input and outputs a scalar \cite{grathwohl_your_2020}. However, as a single energy value is global, we seek to obtain localized energy values corresponding to image patches for segmentation. Hence, we parameterize a \textit{region-based energy model} using a fully convolutional neural network $g_\phi(\hat{S}): \mathbb{R}^{H \times W} \mapsto \mathbb{R}^{K \times K}$ (see Supp. Mat. for details), mapping each patch within a segmentation map $\hat{S}$ to a scalar energy value. This yields $K \times K$ energy map where each patch is $h \times w$ for $h = \frac{H}{K}, w = \frac{W}{K}$. 

Obtaining ground-truth energy values is intractable since it requires marginalizing over the distribution. 
Therefore, we formulate this into a discriminative task where each patch belongs to one of two classes: the positive class denotes patches within the plausible shape distribution, and the negative class, out of distribution. 
Specifically, the class label for each patch $s_s \in \mathbb{R}^{h \times w}$ from $\hat{S}$ is denoted as $y_s \in \{0, 1\}$. 
Assuming an i.i.d. collection of patches from the source data, denoted by $\{s_s^i, y_s^i\}_{i=1}^{N_p} \in \mathcal{D}_s$, with $N_p = N_s \cdot K^2$ as the total number of patches, we formulate the training process using maximum likelihood estimation
\begin{equation}
    \phi^* = \arg \max_\phi \prod_{i=1}^{N_p}P(y_s^i|s_s^i;\phi).
\end{equation}
We minimize the negative log-likelihood (NLL), yielding 
\begin{equation}
    \mathcal{L}_\phi = -\frac{1}{N_p}\log\prod_{i=1}^{N_p}P(y_s^i|s_s^i;\phi).
    \label{eq:nll_loss}
\end{equation}
Categorically, the likelihood of each patch is
\begin{align}
    P(y_s^i|s_s^i;\phi) = 
    \begin{cases}
        \sigma(-g_\phi(s_s^i)), & \text{if}\; y_s^i=1\\
        1 - \sigma(-g_\phi(s_s^i)), & \text{if}\; y_s^i=0 \\
    \end{cases}
    \label{eq:likelihood_function}
\end{align}
where $\sigma(x) = \frac{1}{1 + e^{-x}}$ is the sigmoid activation function. By plugging the likelihood function (\cref{eq:likelihood_function}) into \cref{eq:nll_loss}, we derive the training objective for the energy model as a patchwise binary cross-entropy (BCE) loss:
\begin{align}
    \mathcal{L}_\phi = \frac{1}{N_p} \sum_{i=1}^{N_p} &\Big( -y_s^i \log \sigma(-g_\phi(s_s^i)) \notag \\
    &\quad - (1 - y_s^i) \log (1 - \sigma(-g_\phi(s_s^i))) \Big).
    \label{eq:energy_loss}
\end{align}


However, since we are only given access to the source-domain training data, we do not have any negative examples. To address this challenge, we will simulate negative examples by perturbing the inputs of the segmentation network $f_\theta(\cdot)$, which explores the input data space of the energy model $g_\phi(\cdot)$ by generating various segmentation maps.

\begin{algorithm}[tb]
\caption{\revise{Adversarial perturbation}}\label{alg:adv_training}
\KwData{Source image $I_s$, source model $f_\theta(\cdot)$, ground truth mask $S_s$, perturbation magnitude $\delta$}
\KwResult{Perturbed segmentation $\tilde{S}_s$}
Apply FGSM to generate adversarial noise $\epsilon(I_s,S_s,f_\theta(\cdot),\delta)$ by applying \cref{eq:adv_noise}\;
Perturb the input $\tilde{I}_s=I_s+\epsilon$\;
Perform a forward pass $\tilde{S}_s=f_\theta(\tilde{I}_s)$\;
\end{algorithm}

\noindent\textbf{Curating Negative Examples for the Energy Model.}
Our energy formulation assumes that there exists two collections of examples, one following desired distribution (of shapes) and the other out of the distribution. As we are only given source training dataset and a segmentation network $f_\theta(\cdot)$ pretrained on the same dataset, we do not have access to out-of-distribution (negative) examples. One observation is that input data distribution is not random, as it is constrained by the predictions afforded by the segmentation model $f_\theta(\cdot)$. 

Hence, we propose to explore the data space (segmentation maps, $\hat{S}$) of the energy model $g_\phi(\cdot)$ by probing $f_\theta(\cdot)$ with inputs optimized to simulate undesirable or out-of-distribution examples. This is facilitated by perturbing the input, image $I$, to $f_\theta(\cdot)$ adversarially.
To facilitate the exploration of out-of-distribution examples, we craft adversarial perturbations \revise{(\cref{alg:adv_training})} using the fast gradient sign method (FGSM) \cite{goodfellow_explaining_2015}. During training, we randomly perturb half of each minibatch by computing additive adversarial noise:
\begin{align}
    \epsilon = \delta \operatorname{sign} \left(\nabla_{I_s} \mathcal{L}(f_\theta(I_s), S_s)\right), \label{eq:adv_noise}
\end{align}
where $\delta$ is a scalar controlling the magnitude of the perturbation. We use the Dice loss as our objective function, denoted by $\mathcal{L}$. A forward pass through $f_\theta(\cdot)$ is then performed with the adversarially perturbed image to generate the perturbed segmentation, $\tilde{S_s} = f_\theta(I_s + \epsilon)$.

Adversarial perturbations push the source data manifold into low-density regions where the model has seen few or no examples \cite{goodfellow_explaining_2015}, aligning with OOD definitions as they exhibit lower likelihood under the training distribution \cite{song2017pixeldefend}. By perturbing inputs to induce shifts in segmentation, we expose the energy model to diverse shape variations. To further enhance this, we apply spatial affine transformations and pixel-wise noise with probability $p$ (see Supp. Mat.). After curating perturbations using the above strategy, we compare the perturbed shapes with the ground truth shapes to generate categorical energy labels.




\noindent\textbf{Categorical Energy Label Curation.}
Given the perturbed segmentation maps $\tilde{S}_s$ and the corresponding ground truth $S_s$ from the source dataset, we first divide each input into patches, denoted as $\tilde{s}_s$ and $s_s$. Following the intuition that out-of-distribution or undesired (i.e., perturbed) patches should be assigned a high energy value and labeled as 1, we define the patchwise label $y_s$ as
\begin{align}
    y_s = 1 - \mathbf{1}(d(\tilde{s}_s, s_s) < \tau),
    \label{eq:label_curation}
\end{align}
where $d(\cdot)$ is a distance metric used to measure similarity between patches, and $\tau$ is a threshold for classifying patches as similar. With the perturbation and label curation process defined above, we can now train the region-based energy model using \cref{eq:energy_loss} on the source data. Once trained, the energy model $g_\phi(\cdot)$ will be able to discriminate between desirable and implausible shapes of anatomy. The energy values of $g_\phi(\cdot)$ can be used to update  $f_\theta(\cdot)$ at test time. 

\noindent\textbf{Progressive Adaptation to Match Reference Energy.}
We formulate our test-time objective to align the energy model’s predictions on test samples with a reference (low) energy, represented by a uniform zero matrix $\mathbf{0}_{K \times K}$. Intuitively, our objective aims to update $f_\theta(\cdot)$ such that it reduces the energy predicted by $g_\phi(\cdot)$. To achieve this, we iteratively update the segmentation model as follows:
\begin{align}
    \theta^* = \arg\min_\theta -\sum_{i=1}^{B_t} \log(1 - \sigma(-g_\phi(\hat{s}_t^i))),
    \label{eq:adaptation_loss}
\end{align}
where $\hat{s}_t$ denotes a patch of the segmentation prediction $\hat{S}_t$ on the target dataset and $B_t = K^2 \cdot B$ denotes the number of patches within a minibatch size of $B$. By progressively updating the segmentation model using \cref{eq:adaptation_loss}, we encourage the correction of undesired OOD segmentation predictions, steering them toward realistic shapes.

\input{Tables/mri2d_acdc}

%% file: Tables/mri2d_acdc.tex
\begin{table*}[tb]
\centering
\begin{adjustbox}{scale=0.9}
\footnotesize
\begin{tabular}{llcccc|cccc|ccccc}
\toprule
& & \multicolumn{4}{c|}{ACDC \cite{bernard_deep_2018} $\mapsto$ LVQuant \cite{xue_full_2018}} & \multicolumn{4}{c|}{ACDC \cite{bernard_deep_2018} $\mapsto$ MyoPS \cite{li_myops_2023}} & \multicolumn{4}{c}{ACDC \cite{bernard_deep_2018} $\mapsto$ M\&M \cite{campello_multi-centre_2021}} & \multirow{3}{*}{Avg. Rank}\\
\cmidrule(lr){3-6} \cmidrule(lr){7-10} \cmidrule(lr){11-14}
& & \multicolumn{2}{c}{LV} & \multicolumn{2}{c|}{Myo} & \multicolumn{2}{c}{LV} & \multicolumn{2}{c|}{Myo} & \multicolumn{2}{c}{LV} & \multicolumn{2}{c}{Myo} \\
\cmidrule(lr){3-4} \cmidrule(lr){5-6} \cmidrule(lr){7-8} \cmidrule(lr){9-10} \cmidrule(lr){11-12} \cmidrule(lr){13-14}
& & DSC $\uparrow$ & ASD $\downarrow$ & DSC $\uparrow$ & ASD $\downarrow$ & DSC $\uparrow$ & ASD $\downarrow$ & DSC $\uparrow$ & ASD $\downarrow$ & DSC $\uparrow$ & ASD $\downarrow$ & DSC $\uparrow$ & ASD $\downarrow$ \\
\midrule
\parbox[t]{2mm}{\multirow{5}{*}{\rotatebox[origin=c]{90}{UNet}}}
& Pretrained \cite{ronneberger_u-net_2015} & 58.98 & 24.40 & 42.52 & 19.37 & 85.69 & 2.99 & 72.91 & 2.26 & 47.69 & 24.11 & 41.19 & \underline{15.89} & 4.33\\

& TENT \cite{wang_tent_2020} & 65.78 & \underline{15.37} & 51.57 & 12.78 & 85.63 & \underline{2.94} & 73.49 & 3.24 & 57.01 & \underline{21.15} & \underline{48.26} & 19.99 & \underline{2.92}\\

& CoTTA \cite{wang_continual_2022} & 64.58 & 17.69 & 50.52 & 13.80 & 85.64 & 2.96 & 73.47 & 3.24 & \underline{52.98} & 27.55 & 46.72 & 24.65 & 3.67 \\

& TEA \cite{yuan_tea_2024} & \underline{67.96} & 16.42 & \underline{54.10} & \textbf{11.17} & \underline{85.88} & 3.21 & \underline{73.98} & \underline{2.86} & 52.83 & 38.43 & 48.06 & 29.32 & \underline{2.92}\\ 

& Ours & \textbf{76.93} & \textbf{8.77} & \textbf{59.43} & \underline{11.68} & \textbf{86.06} & \textbf{2.93} & \textbf{78.89} & \textbf{1.91} & \textbf{61.84} & \textbf{19.28} & \textbf{53.13} & \textbf{15.88} & \textbf{1.08}\\
\midrule 

\parbox[t]{2mm}{\multirow{5}{*}{\rotatebox[origin=c]{90}{MedNeXt}}}
& Pretrained \cite{roy_mednext_2023} & 57.55 & 8.67 & 42.26 & 4.80 & 84.39 & 3.39 & 75.77 & 2.07 & 78.43 & 5.48 & 61.06 & 2.95 & 4.67 \\

& TENT \cite{wang_tent_2020} & 75.10 & 6.10 & 54.91 & 3.97 & \underline{84.48} & \underline{3.35} & 75.92 & 2.04 & 83.18 & \underline{4.53} & 67.56 & \underline{2.70} & \underline{2.83}\\

& CoTTA \cite{wang_continual_2022} & 74.57 & 6.32 & 54.85 & 3.93 & 84.46 & 3.36 & \underline{75.95} & \underline{2.03} & 82.90 & 4.83 & \underline{67.93} & 2.89 & 3.25\\

& TEA \cite{yuan_tea_2024} & \underline{75.85} & \underline{5.96} & \underline{55.32} & \underline{3.88} & 84.12 & 3.44 & 75.25 & 2.07 & \underline{83.53} & 4.64 & 67.84 & 2.77 & 3.17 \\ 

& Ours & \textbf{76.22} & \textbf{5.29} & \textbf{57.29} & \textbf{3.70} & \textbf{84.78} & \textbf{3.28} & \textbf{76.44} & \textbf{1.98} & \textbf{83.82} & \textbf{4.11} & \textbf{68.40} & \textbf{2.49} & \textbf{1.00}\\
\midrule 

\parbox[t]{2mm}{\multirow{5}{*}{\rotatebox[origin=c]{90}{SwinUNETR}}}
& Pretrained \cite{hatamizadeh_swin_2022} & 68.44 & \underline{5.92} & 47.64 & 4.20 & 84.84 & 3.26 & 76.35 & 1.99 & 81.92 & \underline{3.52} & 61.83 & \underline{3.03} & 4.25\\

& TENT \cite{wang_tent_2020} & 74.06 & 6.64 & 54.15 & 4.18 & 85.06 & 3.20 & 77.38 & 1.98 & 83.27 & 4.02 & 67.26 & 3.59 & 4.08 \\

& CoTTA \cite{wang_continual_2022} & 73.41 & 6.38 & 54.19 & 4.19 & \underline{85.18} & \underline{3.19} & 77.72 & \underline{1.91} & 83.43 & 3.87 & 67.61 & 3.47 & 2.92\\

& TEA \cite{yuan_tea_2024} & \underline{74.32} & 5.99 & \textbf{54.73} & \underline{4.11} & 85.04 & \underline{3.19} & \underline{77.79} & \underline{1.91} & \textbf{83.93} & 3.90 & \textbf{68.60} & 3.58 & \underline{2.33} \\ 

& Ours & \textbf{76.05} & \textbf{5.79} & \underline{54.22} & \textbf{3.98} & \textbf{85.22} & \textbf{3.17} & \textbf{77.87} & \textbf{1.90} & \underline{83.80} & \textbf{3.13} & \underline{68.15} & \textbf{2.66} & \textbf{1.25}\\
\bottomrule 
\end{tabular}
\end{adjustbox}
\caption{Quantitative comparisons of adapted predictions based on contour-based metrics, using ACDC as the source dataset. Metrics reported include DSC (\%) and ASD (px). The best method is highlighted in bold, and the second best is underlined. 
}
\label{tab:mri2d_acdc}
\end{table*}

%% file: sec/4_experiments.tex
\section{Experiments}
\noindent\textbf{Dataset and Metrics.}
\revise{We evaluated our approach on eight public MRI (bSSFP, T1-, and T2-weighted) and X-ray datasets covering cardiac, spinal cord, and lung segmentation. For cardiac segmentation, we extracted the left ventricle (LV) and myocardium (Myo) and tested on (1) ACDC \cite{bernard_deep_2018}, (2) LVQuant \cite{xue_full_2018}, (3) MyoPS \cite{li_myops_2023}, and (4) M\&M \cite{campello_multi-centre_2021}. For spinal cord segmentation, we used (5) GMSC \cite{prados2017spinal}, which includes samples from four sites (1-4). For lung segmentation, we tested on (6) Shenzhen (CHN) \cite{jaeger2014two}, (7) Montgomery (MCU) \cite{jaeger2014two}, and (8) JSRT \cite{shiraishi2000development}. These datasets exhibit diverse distributions, collected from different hospitals with varying imaging parameters. Patient populations also vary, e.g., ACDC includes five cardiac condition subgroups, whereas MyoPS focuses on myocardial infarction. We preprocess images as follows: cardiac images are resized to $256 \times 256$, spinal cord samples are center-cropped to $144 \times 144$, and lung images are resized to $128 \times 128$. Detailed preprocessing steps are in Supp. Mat.}

We evaluated the performance of our proposed approach by computing the similarity between adapted prediction masks and ground truth masks in terms of (1) Dice coefficient score (DSC) and (2) average surface distance (ASD). Definitions of metrics can be found in the Supp. Mat.

\noindent\textbf{Baselines.}
\revise{Our proposed approach can be plugged and played into existing segmentation frameworks. To test its versatility, we tested on three segmentation networks per cardiac dataset: (1) UNet \cite{ronneberger_u-net_2015}, (2) MedNeXt \cite{roy_mednext_2023} and (3) SwinUNETR \cite{hatamizadeh_swin_2022}. We compared our approach with three test-time adaptation methods for multiclass cardiac segmentation: (1) TENT \cite{wang_tent_2020}, (2) CoTTA \cite{wang_continual_2022}, and (3) TEA \cite{yuan_tea_2024}, integrating them into the segmentation task. For single-class segmentation in spinal cord and lung, we included (4) InTENT \cite{dong_medical_2024} along with additional baselines: (5) SAR \cite{niu_towards_2023}, (6) FSeg \cite{hu_fully_2021}, and (7) MEMO \cite{zhang_memo_2022}.} We update the BatchNorm layers for all methods in accordance with existing conventions \cite{wang_tent_2020, wang_continual_2022, yuan_tea_2024} and restore the model's weights after each batch.

\noindent\textbf{Implementation.}
All our experiments were implemented using PyTorch on NVIDIA A5000 GPUs with 24 GB memory. The architecture of the energy model 
can be found in the Supp. Mat. 
We choose patch size $h=w=16$ and use the mean absolute difference as the distance metric in \cref{eq:label_curation} with $\tau=50$. We use Adam optimizer \cite{kingma_adam_2017} for \cref{eq:adaptation_loss} with 10 iterations. 


%% file: sec/5_results.tex
\section{Results}
\label{sec:results}

\input{Tables/mri2d_mnm}

\input{Tables/gmsc2d}
\input{Tables/chestxray2d}

\noindent\textbf{Main Results.}
We first present a quantitative analysis of test-time adaptation for \revise{cardiac} segmentation in \cref{tab:mri2d_acdc} with ACDC as the source dataset and in \cref{tab:mri2d_mnm} using M\&M as the source. In both tables, our method consistently outperforms baselines across various datasets and architectures, especially in terms of the Dice score, highlighting the effectiveness and robustness of our method. The improvements also show that our trained shape energy model is model-agnostic, as we applied it for three different representative architectures. It is also generalizable as a single energy model and can be applied to a variety of target datasets, e.g., an energy model trained on the ACDC can be agnostically applied to LVQuant, MyoPS, and M\&M for test-time adaptation. \revise{A similar performance trend is also observed in spinal cord (\cref{tab:gmsc}) and lung segmentation (\cref{tab:lung}).}

\begin{figure}[t]
    \centering
    \includegraphics[scale=0.12]{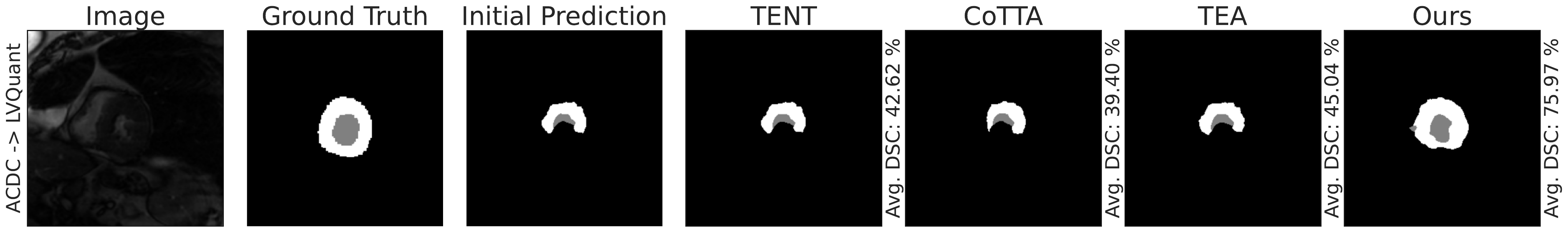} \\
    \includegraphics[scale=0.12]{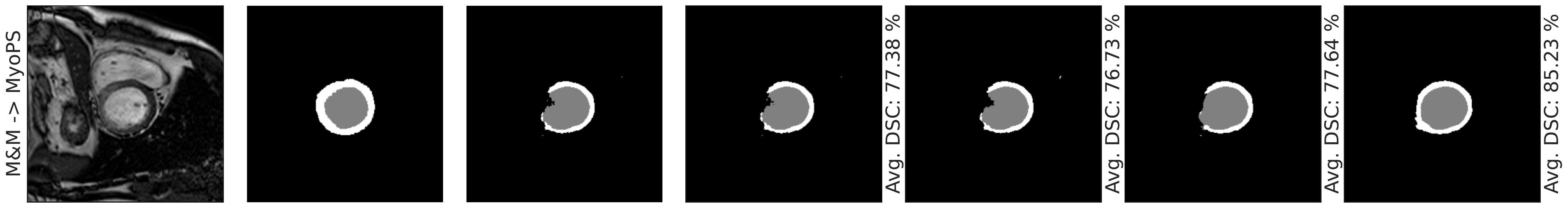} \\
    \includegraphics[scale=0.12]{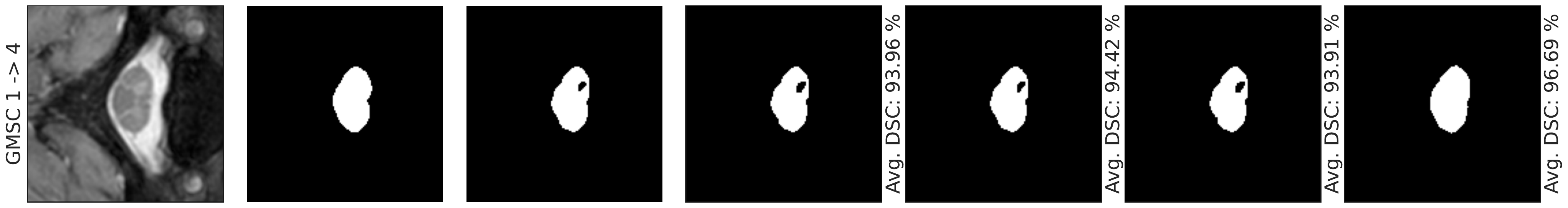} \\
    \includegraphics[scale=0.12]{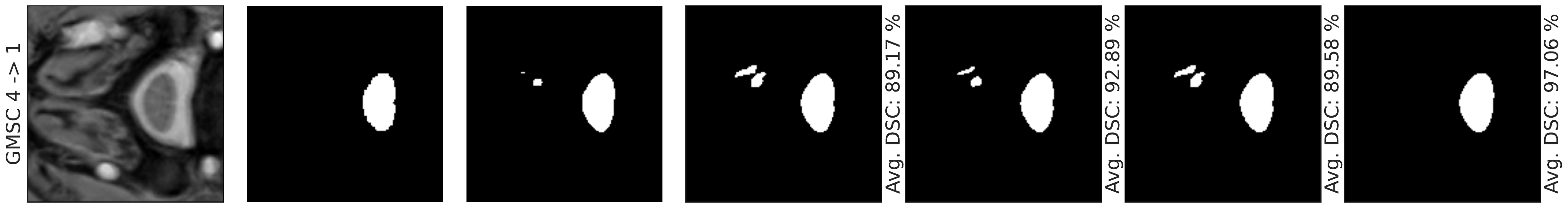} \\
    \includegraphics[scale=0.12]{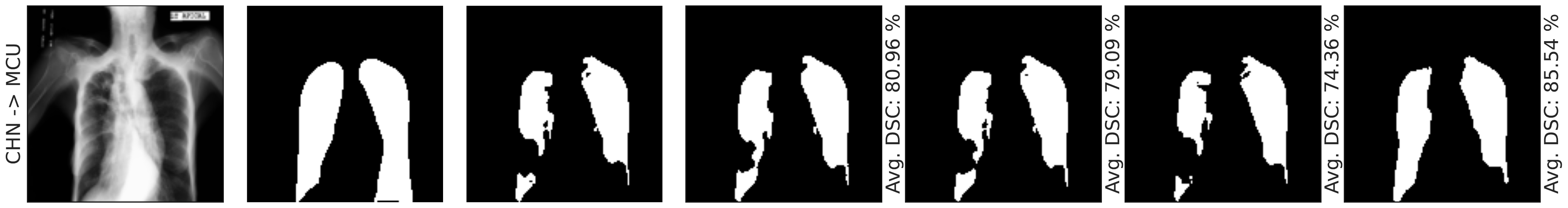} \\
    \includegraphics[scale=0.12]{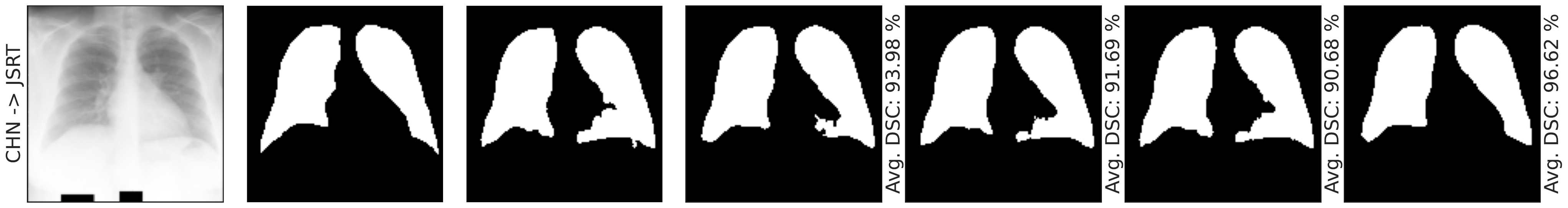}
    \caption{\revise{Qualitative comparison. Rows 1–2 show cardiac segmentation, rows 3–4 spinal cord, and rows 5–6 lung segmentation. Our method refines incomplete initial predictions, producing more plausible shapes after adaptation.} 
    }
    \label{fig:main_results}
\end{figure}

\begin{figure}[tb]
    \centering
    \begin{tabular}{cl}
         \includegraphics[scale=0.15]{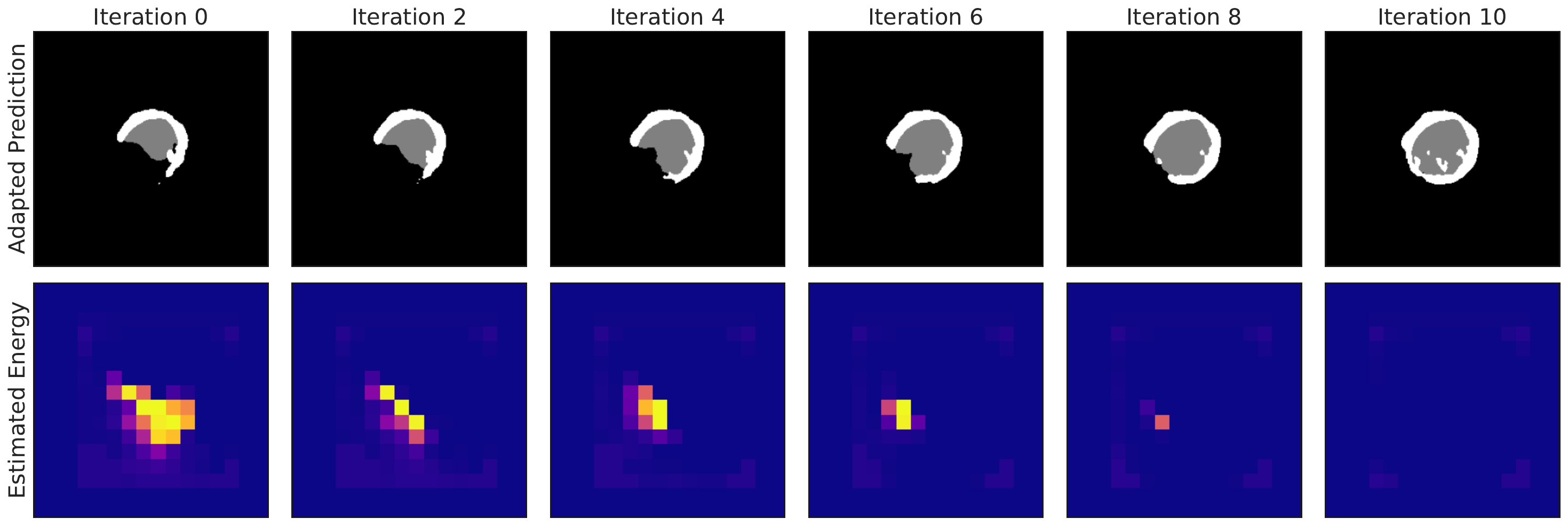} & \hspace{-5mm} \includegraphics[scale=0.2]{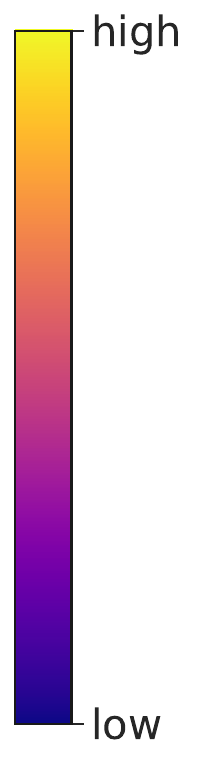}
    \end{tabular}
    \caption{Qualitative evaluation of energy-guided adaptation across iterations. The estimated energy (lower values preferred) in the second row effectively highlights undesired shape regions in the first row. Our approach progressively adapts the source model, refining and completing the initial segmentation in test time.}
    \label{fig:progressive}
\end{figure}

\Cref{fig:main_results} shows these improvements, demonstrating how our method effectively refines and completes erroneous initial predictions from the pretrained model. This is enabled by the progressive updates guided by estimated energy values, and \cref{fig:progressive} illustrates the iterative refinements.
Notably, in the ACDC $\mapsto$ LVQuant, our method improves upon the second best approach by about 9\% for LV and 5\% for Myo in terms of Dice using the UNet. 
\Cref{fig:main_results}, row 1, validates this substantial improvement, showing how our approach successfully completes the left ventricle and myocardium in the initial segmentation map. This is made possible by the implicit shape prior learned by the region-based energy model, which is specifically trained to differentiate between in-distribution and out-of-distribution shapes. Additionally, we consistently yield higher Dice scores across the MedNeXt and SwinUNETR \revise{in cardiac segmentation}, with the sole exception of being comparable to TEA for Myo when using SwinUNETR. We observe that the margin of improvement is smaller when the initial segmentation already achieves sufficiently high Dice, as seen in ACDC $\mapsto$ MyoPS. This is due to that the level of correction needed is relatively small, which is consistent with our design.


\Cref{fig:progressive} illustrates the progression of improvements by showing the estimated energy predictions across iterations. Our region-based energy model accurately identifies regions with undesired shapes by assigning them high energy values in row 2, guiding the model adaptation through iterative computation of \cref{eq:adaptation_loss} to align with the uniform low energy reference. The progressive nature of our update serves as a strength, as gradient updates are local approximations of the loss landscape. Performing a large update may overshoot, which motivates our choice to progressively adapt the model parameters. \revise{This is validated by \cref{tab:mri2d_acdc,tab:mri2d_mnm,tab:gmsc,tab:lung} and qualitative results \cref{fig:main_results}.} Our energy model at test time also provides a coarse level of interpretability in the update as the adaptation is localized to each patch, e.g., when a patch exhibits low energy, the parameters are not updated and errors can be identified visually as energy maps. This further demonstrates the superiority of our method, which directly utilizes an energy-based model to enforce a shape prior by distinguishing in-distribution from out-of-distribution shapes robustly across iterations. This contrasts with existing methods like TEA, which use energy-based models to generate rather than discriminate fake shapes. Consequently, our method is more computationally efficient (see Supp. Mat.) as it does not require generative steps. 

\input{Tables/ablation_module_batch}
\noindent\textbf{Ablation Study.}
\cref{tab:ablation_method} ablates each module. We begin by comparing our region-based formulation (RT) with a global formulation that predicts a single energy value for each segmentation shown in the first row. Our proposed region-based formulation achieves superior performance by providing a more granular estimation, which enhances model adaptation and robustness to background variations. Next, we perform an ablation study on the proposed adversarial perturbations (Adv) and spatial augmentations (SA), demonstrating that our method achieves the best results.

\noindent\textbf{Sensitivity on batch size.}
\revise{To evaluate the effectiveness of our approach with smaller batch sizes, we perform a sensitivity analysis (\cref{tab:gmsc_batchsize}) with batch sizes as low as 1 on GMSC datasets. Our method remains robust and continues to outperform baselines (\cref{tab:gmsc}).}

\noindent\textbf{Complexity analysis.}
\revise{To assess computational complexity, we measure GFLOPs of adversarial perturbation across architectures (UNet: 0.48, MedNeXt: 1.16, SwinUNETR: 1.20) for a single 128$\times$128 sample, showing its cost is comparable to a standard forward pass (UNet: 0.48, MedNeXt: 1.16, SwinUNETR: 1.20). We also analyze time complexity by scaling batch size from 1 to 16 exponentially, recording runtimes of 0.01, 0.23, 0.33, 0.61, and 1.30 seconds. This confirms an $O(N)$ complexity w.r.t. dataset size $N$, demonstrating linear scalability across different architectures.}

\begin{figure}[tb]
    \centering
    \includegraphics[scale=0.16]{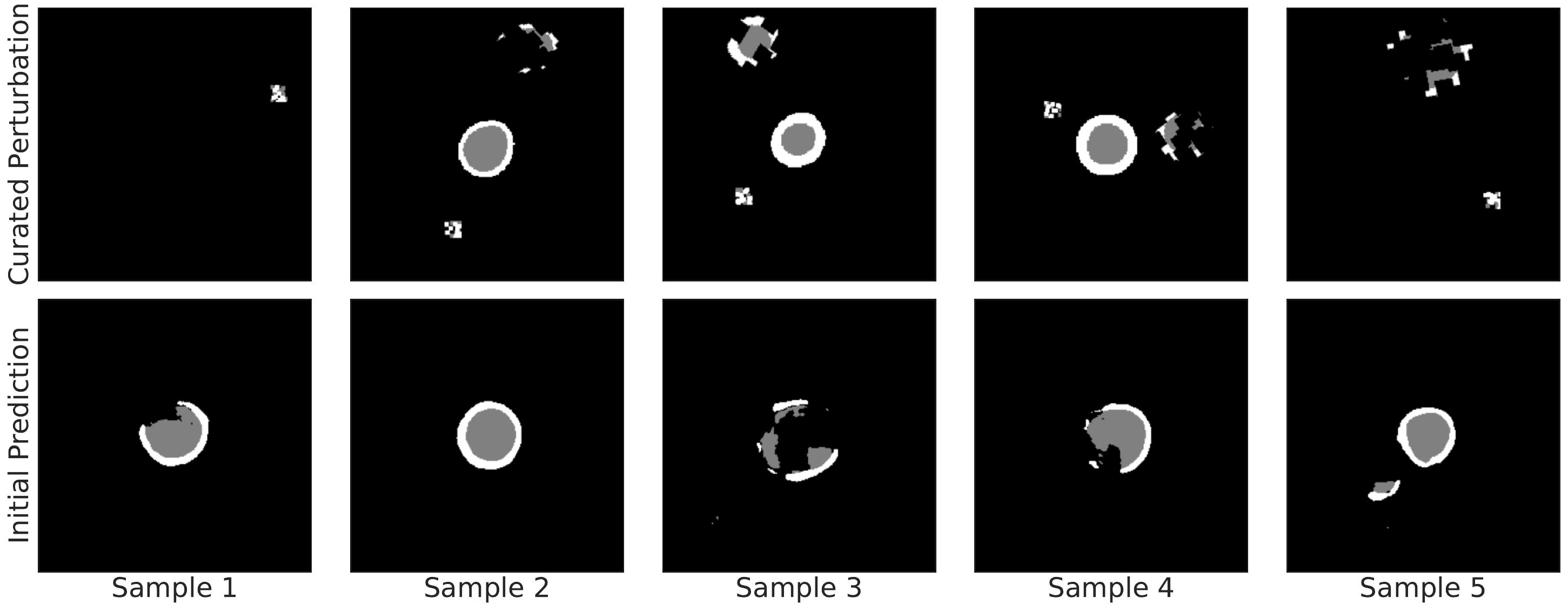}
    \caption{\revise{Qualitative comparison of curated perturbations (top row) and real out-of-distribution (OOD) segmentation errors (bottom row) at test time. The visual similarity between curated and real OOD examples validates the effectiveness of curated perturbations in modeling realistic segmentation errors.}}
    \label{fig:curated_perturbation}
\end{figure}

\input{Tables/ood_validation}
\noindent\textbf{Validation on curated perturbations.}
\revise{
To evaluate our energy model’s ability to detect erroneous regions in OOD segmentation at test time, we compute the accuracy between the estimated energy scores and ground truth labels derived using \cref{eq:label_curation}. This validation serves only for effectiveness analysis, and we do not use them to select the energy model since ground truth in testing is unavailable during adaptation. \cref{tab:ood_validation} shows that our model achieves over 92\% accuracy across architectures, confirming its effectiveness in identifying errors independent of the backbone. \Cref{fig:curated_perturbation} further illustrates the alignment between curated perturbations and true OOD examples, validating our perturbation strategy for modeling real segmentation failures. 
}

\section{Discussion}
\noindent\textbf{Update Model v.s. Update Input.}
Beyond updating the model, one could iteratively refine the initial segmentation map by computing gradients with respect to the input, similar to classical energy-based updates in generative modeling \cite{ho_denoising_2020}. However, it introduces spurious updates that do not align with the segmentation output. Consequently, the updated input would shift to a distribution mismatched with the segmentation network’s learned output space. As a result, it diverges from the distribution shape energy model was trained on, compromising both stability and accuracy.



\noindent\textbf{Limitation.}
\revise{Our approach operates under the common domain adaptation assumption of covariate shift \cite{zhou2021bayesian} and does not explicitly encode image semantics in progressive updates, which may lead to suboptimal performance when source and target distributions differ (e.g., two-chamber vs. four-chamber images). Future work will address this limitation and extend our method to 3D images.}

%% file: Tables/mri2d_mnm.tex
\begin{table*}[tb]
\centering
\begin{adjustbox}{scale=0.9}
\footnotesize
\begin{tabular}{llcccc|cccc|ccccc}
\toprule
& & \multicolumn{4}{c|}{M\&M \cite{campello_multi-centre_2021} $\mapsto$ LVQuant \cite{xue_full_2018}} & \multicolumn{4}{c|}{M\&M \cite{campello_multi-centre_2021} $\mapsto$ MyoPS \cite{li_myops_2023}} & \multicolumn{4}{c}{M\&M \cite{campello_multi-centre_2021} $\mapsto$ ACDC \cite{bernard_deep_2018}}  & \multirow{3}{*}{Avg. Rank}\\
\cmidrule(lr){3-6} \cmidrule(lr){7-10} \cmidrule(lr){11-14}
& & \multicolumn{2}{c}{LV} & \multicolumn{2}{c|}{Myo} & \multicolumn{2}{c}{LV} & \multicolumn{2}{c|}{Myo} & \multicolumn{2}{c}{LV} & \multicolumn{2}{c}{Myo} \\
\cmidrule(lr){3-4} \cmidrule(lr){5-6} \cmidrule(lr){7-8} \cmidrule(lr){9-10} \cmidrule(lr){11-12} \cmidrule(lr){13-14}
& & DSC $\uparrow$ & ASD $\downarrow$ & DSC $\uparrow$ & ASD $\downarrow$ & DSC $\uparrow$ & ASD $\downarrow$ & DSC $\uparrow$ & ASD $\downarrow$ & DSC $\uparrow$ & ASD $\downarrow$ & DSC $\uparrow$ & ASD $\downarrow$ \\
\midrule
\parbox[t]{2mm}{\multirow{5}{*}{\rotatebox[origin=c]{90}{UNet}}}
& Pretrained \cite{ronneberger_u-net_2015} & 89.08 & \underline{2.97} & 75.72 & 2.61 & 75.80 & \underline{5.35} & 57.22 & \textbf{3.22} & 40.84 & 28.45 & 39.10 & 17.28 & 4.08\\

& TENT \cite{wang_tent_2020} & 92.03 & 3.16 & 78.05 & 2.57 & 77.34 & 6.41 & 61.84 & 4.54 & 52.74 & 19.85 & 48.75 & 13.54 & 3.67\\

& CoTTA \cite{wang_continual_2022} & \underline{92.30} & 4.35 & 78.33 & \underline{2.53} & \underline{78.14} & 6.19 & 61.96 & 4.53 & 52.75 & 21.86 & 48.75 & 14.41 & 3.08 \\

& TEA \cite{yuan_tea_2024} & 92.27 & 3.36 & \underline{78.44} & 2.63 & 77.75 & 6.80 & \underline{62.35} & 4.54 & \underline{56.68} & \underline{17.51} & \underline{51.81} & \underline{13.26} & \underline{3.00}\\ 

& Ours & \textbf{93.25} & \textbf{1.57} & \textbf{81.34} & \textbf{1.49} & \textbf{79.14} & \textbf{5.18} & \textbf{68.20} & \underline{3.62} & \textbf{59.97} & \textbf{16.15} & \textbf{54.28} & \textbf{10.68} & \textbf{1.08}\\
\midrule 

\parbox[t]{2mm}{\multirow{5}{*}{\rotatebox[origin=c]{90}{MedNeXt}}}
& Pretrained \cite{roy_mednext_2023} & 92.65 & 1.64 & 79.84 & \underline{1.85} & 79.36 & 4.80 & 67.10 & \underline{2.37} & \underline{87.60} & 4.34 & 73.57 & 3.48 & 3.83\\

& TENT \cite{wang_tent_2020} & 92.92 & 1.56 & 79.90 & 2.19 & \underline{79.78} & 4.68 & \textbf{67.92} & 2.40 & 87.52 & \textbf{3.48} & \underline{74.46} & \textbf{2.76} & \underline{2.75}\\

& CoTTA \cite{wang_continual_2022} & 92.94 & 1.56 & 79.92 & 2.18 & 79.77 & \underline{4.67} & \underline{67.90} & 2.38 & 84.73 & 5.25 & 71.25 & 4.52 & 3.42 \\

& TEA \cite{yuan_tea_2024} & \underline{92.99} & \underline{1.55} & \underline{80.11} & 2.16 & 79.74 & 4.69 & 67.89 & 2.42 & 84.00 & 4.53 & 67.54 & 3.34 & 3.33\\ 

& Ours & \textbf{93.20} & \textbf{1.51} & \textbf{81.41} & \textbf{1.70} & \textbf{80.09} & \textbf{4.58} & 67.87 & \textbf{2.32} & \textbf{88.09} & \underline{4.25} & \textbf{74.99} & \underline{3.27} & \textbf{1.42}\\
\midrule 

\parbox[t]{2mm}{\multirow{5}{*}{\rotatebox[origin=c]{90}{SwinUNETR}}}
& Pretrained \cite{hatamizadeh_swin_2022} & 92.98 & 2.40 & 80.68 & 1.66 & 81.30 & 4.27 & 71.84 & \underline{2.29} & \underline{88.49} & \textbf{2.66} & \underline{77.84} & 2.43 & 3.75\\

& TENT \cite{wang_tent_2020} & 93.13 & 1.48 & 81.22 & 1.64 & 81.43 & 4.24 & 72.19 & 2.34 & 88.27 & \underline{2.74} & 77.49 & \underline{2.37} & \underline{2.92}\\

& CoTTA \cite{wang_continual_2022} & 93.11 & 1.51 & 81.10 & 1.64 & \underline{81.48} & \underline{4.23} & \underline{72.27} & 2.33 & 88.10 & 2.90 & 77.41 & 2.40 & 3.25\\

& TEA \cite{yuan_tea_2024} & \underline{93.20} & \underline{1.47} & \textbf{81.97} & \underline{1.58} & 81.20 & 4.30 & 71.90 & 2.35 & 87.89 & 3.54 & 77.07 & \underline{2.37} & 3.58\\ 

& Ours & \textbf{93.31} & \textbf{1.45} & \underline{81.17} & \textbf{1.53} & \textbf{81.71} & \textbf{4.21} & \textbf{72.45} & \textbf{2.17} & \textbf{88.52} & 3.04 & \textbf{78.15} & \textbf{2.24} & \textbf{1.33}\\
\bottomrule 
\end{tabular}
\end{adjustbox}
\caption{Quantitative comparisons of adapted predictions based on contour-based metrics, using M\&M as the source dataset. Metrics reported include DSC (\%) and ASD (px). The best method is highlighted in bold, and the second best is underlined. 
}
\label{tab:mri2d_mnm}
\vspace{-2mm}
\end{table*}

%% file: Tables/gmsc2d.tex
\begin{table}[tb]
\centering
\begin{adjustbox}{scale=0.8}
\footnotesize
\begin{tabular}{lccccccc}
\toprule
& 1 $\mapsto$ 2 & 1 $\mapsto$ 3 & 1 $\mapsto$ 4 & 4 $\mapsto$ 1 & 4 $\mapsto$ 2 & 4 $\mapsto$ 3 & Avg. \\
\midrule
TENT \cite{wang_tent_2020} & 70.5 & 16.8 & 57.4 & 87.0 & 67.9 & 72.9 & 62.1 \\
SAR \cite{niu_towards_2023} & 72.1 & 17.5 & 59.9 & 85.1 & 66.6 & 72.7 & 62.3 \\
FSeg \cite{hu_fully_2021} & 70.5 & 16.9 & 57.4 & 87.0 & 67.9 & 72.7 & 62.1 \\
MEMO \cite{zhang_memo_2022} & 69.9 & 17.0 & 56.4 & 86.8 & 67.0 & 72.5 & 61.6 \\
CoTTA \cite{wang_continual_2022} & 66.1 & 63.3 & 92.1 & 95.0 & 54.7 & 86.7 & 76.4\\
TEA \cite{yuan_tea_2024} & 68.4 & 66.5 & 92.4 & 94.9 & 54.7 & 86.7 & 77.3\\
InTENT \cite{dong_medical_2024} & \textbf{86.6} & 28.7 & 71.4 & 83.3 & \textbf{79.2} & 75.0 & 70.7 \\
Ours & 73.6 & \textbf{77.7} & \textbf{95.3} & \textbf{95.1} & 56.2 & \textbf{87.2} & \textbf{80.9} \\
\bottomrule 
\end{tabular}
\end{adjustbox}
\caption{\revise{Quantitative comparisons of adapted predictions for spinal cord MRI segmentation, with sites 1 and 4 in GMSC as the source dataset. Reported metrics include DSC (\%).}}
\label{tab:gmsc}
\end{table}

%% file: Tables/chestxray2d.tex
\begin{table}[tb]
\centering
\begin{adjustbox}{scale=0.8}
\footnotesize
\begin{tabular}{lccc}
\toprule
& CHN $\mapsto$ MCU & CHN $\mapsto$ JSRT & Avg.\\
\midrule
TENT \cite{wang_tent_2020} & 86.2 & 95.2 & 90.7\\
SAR \cite{niu_towards_2023} & 85.5 & 95.0 & 90.3 \\
FSeg \cite{hu_fully_2021} & 86.2 & 95.2 & 90.7\\
MEMO \cite{zhang_memo_2022} & 85.0 & 95.1 & 90.1\\
CoTTA \cite{wang_continual_2022} & 95.8 & 95.2 & 95.5\\
TEA \cite{yuan_tea_2024} & 95.7 & 95.5 & 95.6\\
InTENT \cite{dong_medical_2024} & 95.5 & \textbf{96.3} & 95.9\\
Ours & \textbf{96.1} & \textbf{96.3} & \textbf{96.2}\\
\bottomrule 
\end{tabular}
\end{adjustbox}
\caption{\revise{Quantitative comparisons of adapted predictions for chest X-ray lung segmentation, with CHN as the source dataset. Reported metrics include DSC (\%).}}
\label{tab:lung}
\end{table}

%% file: Tables/ablation_module_batch.tex
\begin{table}[tb]
    \centering
    \begin{minipage}{0.24\textwidth}
        \centering
        \begin{adjustbox}{scale=0.8}
        \footnotesize
        \begin{tabular}{lllcc}
            \toprule
            RT & Adv & SA & LV & Myo \\
            \midrule
            \xmark & \cmark & \cmark & 71.03  & 54.17 \\\midrule
            \cmark & \xmark & \cmark & 64.72  & 47.47 \\
            \cmark & \cmark & \xmark & 76.15  & 58.32 \\
            \cmark & \cmark & \cmark & \textbf{76.93}  & \textbf{59.43} \\
            \bottomrule
        \end{tabular}
        \end{adjustbox}
        \caption{\revise{Ablation study of our modules on UNet for ACDC $\mapsto$ LVQuant. Metrics: DSC (\%).}}
        \label{tab:ablation_method}
    \end{minipage}
    \hfill
    \begin{minipage}{0.22\textwidth}
        \centering
        \begin{adjustbox}{scale=0.84}
        \footnotesize
        \begin{tabular}{cccc}
            \toprule
            & 1 $\mapsto$ 2 & 1 $\mapsto$ 3 & 1 $\mapsto$ 4\\
            \midrule
            bs=1 & 75.0 & 83.5 & 95.1\\
            bs=2 & 71.3 & 78.4 & 95.1\\
            bs=3 & 70.3 & 77.6 & 95.5\\
            bs=4 & 73.6 & 77.7 & 95.3\\
            \bottomrule 
        \end{tabular}
        \end{adjustbox}
        \caption{\revise{Sensitivity analysis of batch size on GMSC datasets. Metrics: DSC (\%).}}
        \label{tab:gmsc_batchsize}
    \end{minipage}
\end{table}

%% file: Tables/ood_validation.tex
\begin{table}[tb]
    \centering
    \begin{adjustbox}{scale=0.8}
    \footnotesize
    \begin{tabular}{cccccc}
    \toprule
        Method  & ACDC $\mapsto$ LVQuant & ACDC $\mapsto$ MyoPS & ACDC $\mapsto$ M\&M  \\
        \midrule
        UNet  & 93.64 & 96.55 & 94.53\\
        MedNeXt & 92.17 & 95.83 & 93.66\\
        SwinUNETR & 92.01 & 96.42 & 93.97\\
        \bottomrule
    \end{tabular}
    \end{adjustbox}
    \caption{\revise{Quantitative evaluation of the region-based shape energy model for detecting erroneous regions at test time. We report accuracy (\%) across different architectures.}}
    \label{tab:ood_validation}
\end{table}

%% file: sec/6_conclusion.tex
\section{Conclusion}
\label{sec:conclusion}
\revise{We introduce progressive test-time energy adaptation for medical image segmentation, allowing pretrained models to dynamically adapt to test data. Our method identifies incorrect predictions by assessing whether shapes conform to a plausible anatomical distribution. To achieve this, we propose a region-based energy model, which assigns high energy to erroneous regions and low energy to accurate ones. 
Evaluated on eight public MRI and X-ray datasets covering cardiac, spinal cord, and lung segmentation, our approach consistently outperforms existing methods. Additionally, our model-agnostic design allows seamless integration with existing segmentation networks, making it more practical for real-world applications.}

%% file: sec/X_suppl.tex
\clearpage
\setcounter{page}{1}
\setcounter{section}{0}
\renewcommand{\thesection}{\Alph{section}}
\maketitlesupplementary
\section{Dataset Details}

\subsection{ACDC \cite{bernard_deep_2018}}
The ACDC dataset, publicly accessible, comprises 2D cardiac MRI scans from 150 patients across five subgroups: (1) 30 normal patients, (2) 30 with previous myocardial infarction, (3) 30 with dilated cardiomyopathy, (4) 30 with hypertrophic cardiomyopathy, and (5) 30 with abnormal right ventricles. The acquisitions were obtained using two MRI scanners of different magnetic strengths (1.5 T and 3.0 T). Cine images were acquired in breath hold with an SSFP sequence in short-axis orientation. The spatial resolution ranges from 1.37 to 1.68 $\text{mm}^2$. The dataset includes 100 training and 50 testing subjects. Each sequence has end-diastole (ED) and end-systole (ES) frames with left ventricle and myocardium labels. Our training set includes 80 patients, the validation set 20 patients, and the testing set 50 patients. ED and ES image pairs are extracted slice-by-slice from 2D longitudinal stacks, center-cropped to $256\times 256$ around the myocardium centroid in ED frames after resampling the pixel spacing to $1 \text{mm}\times 1\text{mm}$. This process produces 1266 2D training images, 277 validation images, and 852 testing images. To reduce computational cost, approximately one-quarter of the testing set is randomly subsampled to 213 samples.

\subsection{LVQuant \cite{xue_full_2018}}
The LVQuant dataset is publicly available and includes short-axis MR sequences from 56 subjects. The 2D cine MR images were collected from three hospitals during routine clinical practice without specific selection criteria, encompassing pathologies ranging from moderate to severe cardiac conditions. Each sequence contains 20 frames of mid-ventricle slices spanning a complete cardiac cycle, acquired with ECG-gating and breath-holding. Ground truth segmentations for the endocardium and epicardium are provided. The MR images have pixel spacings ranging from 0.6 mm to 2.08 mm. Following the preprocessing steps of ACDC, we resample the pixel spacings to $1 \text{mm}\times 1 \text{mm}$ and apply a center crop to $256 \times 256$. End-diastole (ED) and end-systole (ES) frames are extracted from each sequence, and myocardium masks are generated by subtracting the endocardium mask from the epicardium. The dataset is randomly split into 60/20/20 for training, validation, and testing, resulting in 66 training images, 22 validation images, and 24 testing images. Due to the limited size of the training set, it is insufficient for developing a robust source segmentation model. Therefore, only the testing set from LVQuant is reserved for adaptation experiments.

\subsection{MyoPS \cite{li_myops_2023}}
The MyoPS challenge dataset includes 45 paired three-sequence CMR images (bSSFP, LGE, and T2 CMR) acquired from the same patients, with 25 subjects publicly available. Each subject contains 2–6 slices, with an in-plane resolution ranging from 0.73 to 0.76 mm. The dataset provides gold-standard segmentations for the left ventricular blood pool, right ventricular blood pool, left ventricular myocardium, left ventricular myocardial scar, and edema. We randomly split the dataset into training, validation, and testing sets in a 60/20/20 ratio, resulting in 51 training images, 23 validation images, and 18 testing images. However, similar to the LVQuant dataset, the limited number of training examples makes it insufficient for developing a source segmentation model. Consequently, we only retain the testing set for adaptation experiments. For these experiments, we extract the bSSFP sequence as the input image and compose the ground truth myocardium by combining the myocardial scar and edema labels with the normal myocardium.

\subsection{M\&M \cite{campello_multi-centre_2021}}
The M\&M dataset is publicly available, comprising 360 cases collected from four vendors (Siemens, Philips, GE Healthcare, and Canon), six centers, and three countries (Spain, Canada, and Germany). The dataset builds upon the labeling framework of the ACDC dataset \cite{bernard_deep_2018}, providing ground truth annotations for the left ventricle, myocardium, and right ventricle. The in-plane resolution ranges from 0.98 to 1.32 mm, and the longitudinal stack contains 10-12 slices. Following a preprocessing pipeline similar to ACDC, we extract the end-diastolic (ED) and end-systolic (ES) frames from each sequence slice-by-slice along the longitudinal stack. After resampling pixel spacings to $1 \text{mm} \times 1 \text{mm}$, we center-crop the frames to $256 \times 256$ around the myocardium mask in the ED frame. The dataset is randomly divided into training, validation, and testing sets in a 60/20/20 ratio, resulting in 2918 training images, 964 validation images, and 987 testing images. To reduce computational costs, the testing set is subsampled to approximately one-quarter of its size, yielding 246 samples.

\subsection{GMSC \cite{prados2017spinal}}
\revise{The GMSC dataset is a multi-center, multi-vendor collection of spinal cord MRI anatomical images, comprising healthy subjects from four sites: (1) Site 1 from University College London, acquired using a 3T Philips Achieva T1-weighted MRI; (2) Site 2 from Polytechnique Montreal, using a 3T Siemens TIM Trio with T1-weighted; (3) Site 3 from the University of Zurich, using a 3T Siemens Skyra T2-weighted MRI; and (4) Site 4 from Vanderbilt University, acquired with a 3T whole-body Philips scanner for T2-weighted. Each site contains 10 subjects, with manual annotations from four raters. Following \cite{dong_medical_2024}, we preprocess the data by center-cropping each slice to $144\times144$ after normalizing image intensity to $[0,1]$ and randomly split it at the subject level into training, validation, and testing sets using a 60/20/20 ratio, resulting in 21/3/6 for Site 1, 76/13/24 for Site 2, 125/18/36 for Site 3, and 95/12/26 for Site 4. This structured distribution ensures balanced evaluation across imaging centers and scanner variations.}

\subsection{CHN \cite{jaeger2014two}}
\revise{The Shenzhen (CHN) chest X-ray dataset, created by the Third People's Hospital of Shenzhen City and Guangdong Medical College in collaboration with the Department of Health and Human Services, is publicly available. It consists of 566 chest X-ray images, including both normal and abnormal cases with tuberculosis manifestations, accompanied by radiologist readings. We preprocess the images by resizing them to $128 \times 128$ and randomly splitting the dataset into training, validation, and testing sets using a 60/20/20 ratio, resulting in 339/113/114 images, respectively.}

\subsection{MCU \cite{jaeger2014two}}
\revise{The Montgomery County dataset (MCU) is publicly available and was created through a collaboration between the National Library of Medicine and the Montgomery County Department of Health and Human Services. It comprises 138 chest X-ray images, including 80 normal cases and 58 with tuberculosis-related abnormalities. Following the same preprocessing as CHN, we resize each image to $128 \times 128$ and randomly split the dataset into training, validation, and testing sets using a 60/20/20 ratio, yielding 82/28/28 images, respectively.}

\subsection{JSRT \cite{shiraishi2000development}}
\revise{The Japanese Society of Radiological Technology (JSRT) dataset is a publicly available collection of posteroanterior chest X-ray (CXR) images, widely used for lung segmentation and nodule detection research. It comprises 199 images for training and 40 for testing. Following standard protocols from other lung segmentation datasets, we randomly split the dataset into training, validation, and testing sets using a 60/20/20 ratio, yielding 159/40/48 images, respectively. Each image is resized to $128\times 128$. This structured split ensures a balanced evaluation across different dataset partitions, facilitating robust model training and validation.}

\section{Implementation Details}
\subsection{Evaluation Metrics}
\paragraph{Dice Score}
For a pair of predicted segmentation mask $\hat{S}\in\{0,1\}^{H\times W}$ and ground truth mask $S\in\{0,1\}^{H\times W}$, the Dice score is defined to measure the ratio of overlap:
\begin{align}
    \text{Dice}(\hat{S}, S) = \frac{2 |\hat{S} \cap S|}{|\hat{S}| + |S|}.
\end{align}
Here, $|\hat{S} \cap S|$ denotes the number of overlapping elements between the predicted mask $\hat{S}$ and the ground truth mask $S$, while $|\hat{S}|$ and $|S|$ represent the total number of elements in the predicted mask and ground truth mask, respectively. The Dice score quantifies the similarity between the predicted and ground truth masks, ranging from 0 to 1. A Dice score of 1 indicates perfect overlap, while a score of 0 indicates no overlap. 

\paragraph{Average Surface Distance}
We compute the Average Surface Distance (ASD) to measure the mean boundary deviation between the predicted and ground truth segmentation masks. Formally, ASD is defined as:
\begin{align}
    d_{\text{ASD}}(\hat{S}, S) = \frac{1}{|\partial \hat{S}| + |\partial S|} 
    \Bigg( \sum_{x \in \partial \hat{S}} \min_{y \in \partial S} \|x - y\|  \Bigg. \\
    \Bigg. + \sum_{y \in \partial S} \min_{x \in \partial \hat{S}} \|y - x\| \Bigg).
\end{align}
Here, \( \partial \hat{S} \) and \( \partial S \) represent the boundary points of the predicted segmentation and the ground truth segmentation, respectively. The term \( \|x - y\| \) denotes the Euclidean distance between two points, while \( \min_{y \in \partial S} \|x - y\| \) computes the shortest distance from a boundary point \( x \in \partial \hat{S} \) to the closest point in \( \partial S \), ensuring an accurate local correspondence. The final ASD value represents the mean of these shortest distances, summing over both segmentations.


\subsection{Training Details}
\paragraph{Source Segmentation Model}
All experiments are implemented in Python using the PyTorch framework. We train the segmentation models from scratch on the source datasets ACDC and M\&M for each architecture, utilizing a hybrid segmentation loss combining Dice and cross-entropy. During training, we apply random data augmentation with a probability of 
0.5, including random flipping, translation, and rotation. The models are optimized using the Adam optimizer with a learning rate of $1\times 10^{-4}$. To align with the convention of adapting BatchNorm layers \cite{wang_tent_2020, wang_continual_2022, yuan_tea_2024} in segmentation models, we replace GroupNorm in MedNeXt and InstanceNorm in SwinUNETR with BatchNorm. All segmentation models are trained with 150 epochs with a batch size of 8. We want to emphasize that our test-time adaptation framework operates under the assumption that the source segmentation model is pre-trained and provided.

\paragraph{Shape Energy Model}
Our region-based shape energy model is implemented as a simple convolutional neural network (CNN), designed to capture the localized nature of shape energy. The model comprises four convolutional layers, each with a kernel size of 5, stride of 2, and padding of 2. Each convolutional layer is followed by a LeakyReLU activation function with a negative slope of 0.2, as well as a BatchNorm layer for regularization and stability. Finally, the output is projected to a single channel using an additional convolutional layer as logits. 
For spatial augmentations, we introduce handcrafted spatial affine transformation and pixel-wise noise with probability $p$ across all samples in the minibatch. Additionally, we apply patchwise dropout to create holes in the initially augmented masks. These altered predictions are then added back to the original masks for further augmentation. We use the \texttt{BCEWithLogitsLoss} to exploit the logsumexp trick for training numerical stability. We use one-hot encoding for the ground truth mask and apply softmax activation for segmentation prediction logits as inputs for the shape energy model. We train our proposed region-based shape energy model using the Adam optimizer with 150 epochs. We use the cosine decay learning rate scheduler with a warm-up stage including 1000 steps. During the adaptation stage, we employ the Adam optimizer to update the collected BatchNorm parameters from the source segmentation model.

\section{Additional Results}
\begin{figure*}[tb]
    \centering
    \includegraphics[scale=0.25]{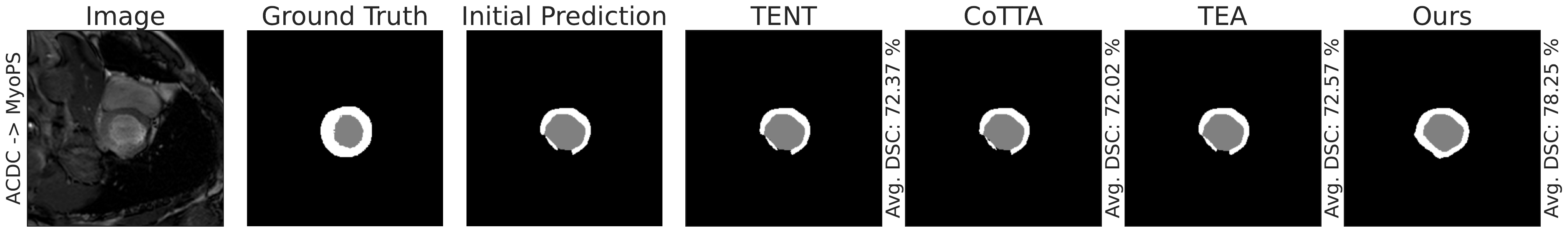}\\
    \includegraphics[scale=0.25]{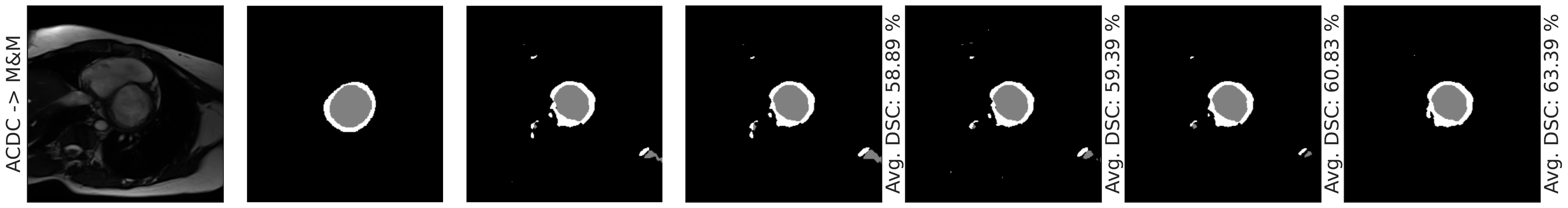} \\
    \includegraphics[scale=0.25]{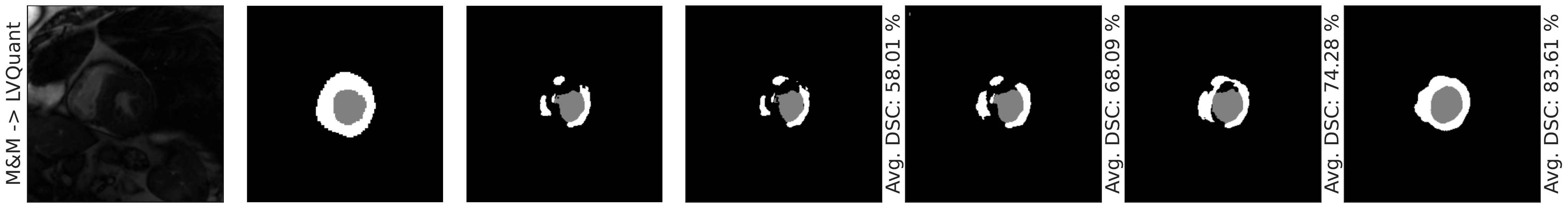}\\
    \includegraphics[scale=0.25]{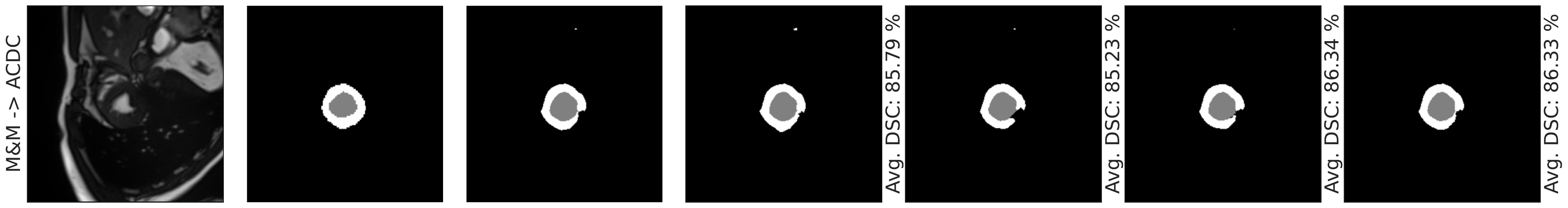} \\
    \includegraphics[scale=0.25]{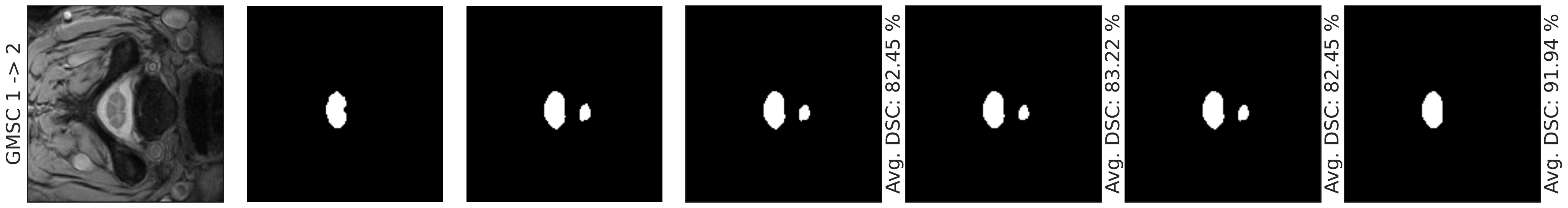}\\
    \includegraphics[scale=0.25]{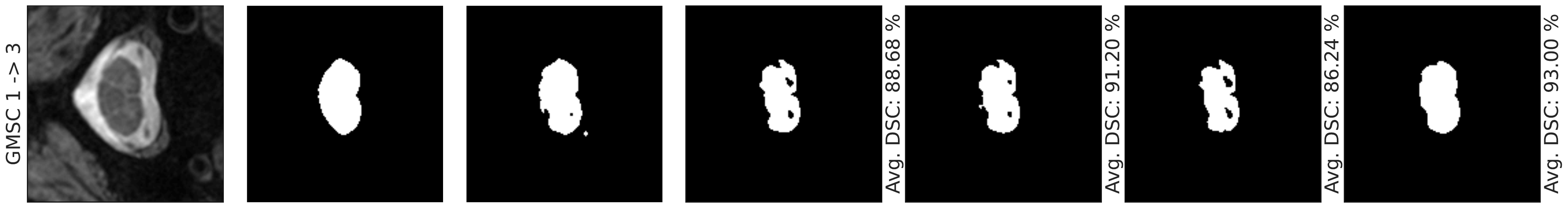}\\
    \includegraphics[scale=0.25]{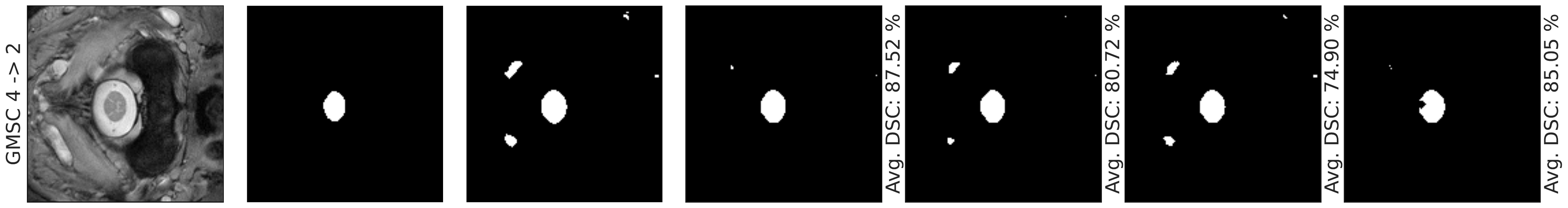}\\
    \includegraphics[scale=0.25]{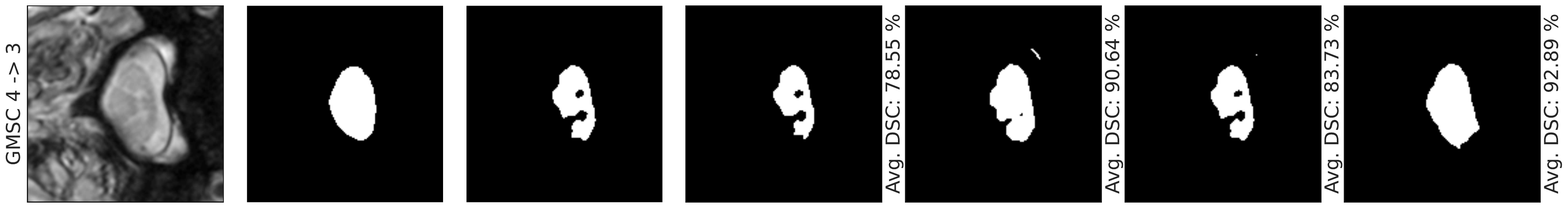}\\
    \caption{\revise{Qualitative evaluation of our proposed approach using the UNet architecture compared to baseline methods. The top four rows depict adapted cardiac segmentation, while the bottom four rows show lung segmentation from chest X-rays. Our approach effectively refines incomplete initial segmentations, generating more anatomically plausible shapes after adaptation.}
    }
    \label{fig:additional_results}
\end{figure*}

\begin{figure*}[tb]
    \centering
    \includegraphics[scale=0.25]{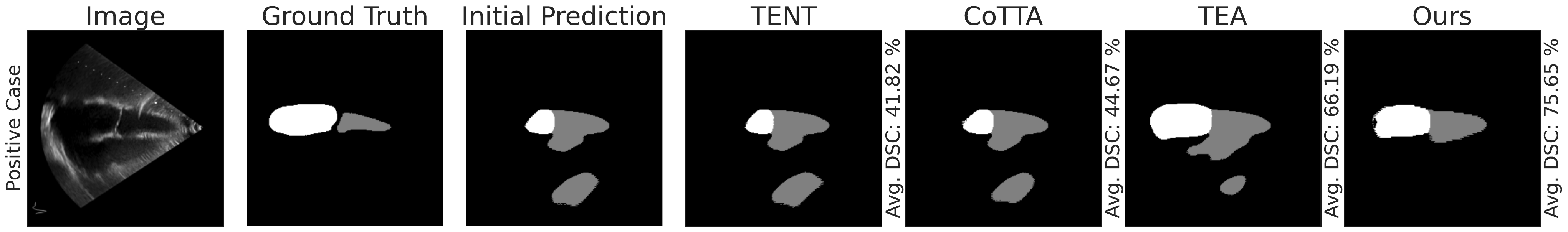} \\
    \includegraphics[scale=0.25]{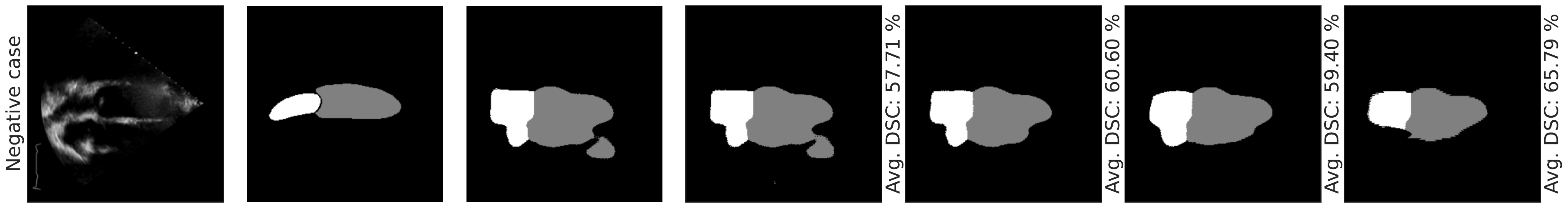}
    \caption{Qualitative evaluation of adaptation performance on source and target datasets with misaligned semantics is presented. Our proposed approach is trained on a 2D ultrasound dataset from CAMUS and adapted to the CardiacUDA Site G dataset. In the top row, we showcase a positive example where the initial prediction accurately identifies the right chambers. Conversely, the bottom row illustrates a negative case where the initial prediction incorrectly identifies the chamber locations. These examples highlight the challenges of semantic misalignment and the variability in adaptation outcomes. 
    }
    \label{fig:camus_to_siteG_example}
\end{figure*}


\input{Tables/ablation_batchnorm}

\input{Tables/ablation_iteration}

\paragraph{Quantitative Results}
We begin by presenting the quantitative evaluation of the source segmentation models, as shown \cref{tab:source_seg_model}. All models demonstrate strong performance in LV segmentation, achieving Dice scores above 90\%, and myocardium segmentation, with Dice scores exceeding 80\%. Myocardium segmentation is inherently more challenging due to its complex structure, being a thin, crescent-shaped layer surrounding the LV. Both MedNeXt and SwinUNETR achieve lower average surface distances compared to the vanilla UNet, highlighting the advantages of their advanced architectures. MedNeXt benefits from its ConvNeXt backbone, while SwinUNETR leverages transformer layers, both of which excel at capturing global context. This enables them to handle fine details more effectively and produce smoother, more accurate contours, particularly for complex structures like the myocardium.

We further analyze the effect of the number of iterations in the proposed method, as shown in \cref{tab:ablation_iteration}. The results indicate that the Dice score improves consistently with an increasing number of iterations, with multiple iterations significantly outperforming a single update. This improvement occurs because a single gradient-based update represents only a linear step, which is insufficient to reach the typically nonlinear local minima required for optimal performance. We observe that the average surface distance for the myocardium increases as the number of iterations grows. This can be attributed to a limitation in our proposed approach, which lacks explicit regularization of the image during adaptation. Consequently, this may lead to the generation of shapes in unintended regions, ultimately contributing to the observed increase in the average surface distance.

\paragraph{Qualitative Results}
We present additional qualitative results of our proposed approach in \cref{fig:additional_results}, demonstrating its ability to effectively refine initial predictions and produce more plausible shapes after adaptation. The refined shapes are visually closer to the ground truth compared to other baselines, showcasing the effectiveness of our method. To further evaluate the performance of our approach in scenarios where imaging semantics are misaligned, we conduct an additional analysis by training on the public 2D ultrasound dataset CAMUS, which contains two-chamber views, and adapting to the CardiacUDA Site G dataset, which consists of four-chamber views. The qualitative results are shown in \cref{fig:camus_to_siteG_example}. In the top row example, when the initial prediction correctly identifies the chamber, our proposed approach successfully removes outliers and generates more plausible shapes compared to the baselines. However, in cases where the initial prediction incorrectly identifies the chamber (bottom row example), our approach fails to correct the misprediction. We plan to address this limitation in future work by introducing enhanced regularization on the image during the adaptation process.

\paragraph{Computation Efficiency}
We evaluate the running time of our proposed approach and compare it with existing methods in \cref{tab:running_time}. Measured on a single GPU during inference, our approach achieves an average speedup of 4.5$\times$ over CoTTA and 1.3$\times$ over TEA, while remaining comparable to TENT. This efficiency stems from our trained energy model, which requires only a forward pass, eliminating the need for extensive augmentation averaging in CoTTA or the stochastic gradient Langevin dynamics used to generate synthetic samples in TEA.
\input{Tables/running_time}



%% file: Tables/ablation_batchnorm.tex
\begin{table}[tb]
    \centering
    \begin{adjustbox}{scale=0.65}
    \begin{tabular}{lcccc|cccc}
    \toprule 
    \multirow{3}{*}{Architecture} & \multicolumn{4}{c|}{ACDC}  & \multicolumn{4}{c}{M\&M} \\
    \cmidrule(lr){2-5} \cmidrule(lr){6-9} 
    & \multicolumn{2}{c}{LV} & \multicolumn{2}{c|}{Myo} & \multicolumn{2}{c}{LV} & \multicolumn{2}{c}{Myo} \\
    \cmidrule(lr){2-3} \cmidrule(lr){4-5} \cmidrule(lr){6-7} \cmidrule(lr){8-9} 
    & DSC $\uparrow$ & ASD $\downarrow$  & DSC $\uparrow$ & ASD $\downarrow$ & DSC $\uparrow$ & ASD $\downarrow$  & DSC $\uparrow$ & ASD $\downarrow$ \\\midrule
    UNet & 90.16 & 3.81 & 82.56 & 4.21 & 92.99 & 2.75 & 84.45 & 2.38\\
    MedNeXt & 90.06 & 1.63 & 80.62 & 1.64 & 94.00 & 1.13 & 84.29 & 1.25 \\
    SwinUNETR & 91.95 & 1.36 & 84.39 & 1.38 & 94.41 & 1.19 & 85.25 & 1.24 \\
    \bottomrule 
    \end{tabular}
    \end{adjustbox}
    \caption{Quantitative results for the source segmentation models (UNet, MedNeXt, and SwinUNETR) trained on the ACDC and M\&M datasets are presented. The evaluation metrics include the DSC (\%) and ASD (px).}
    \label{tab:source_seg_model}
\end{table}

%% file: Tables/ablation_iteration.tex
\begin{table}[tb]
    \centering
    \begin{adjustbox}{scale=0.8}
    \begin{tabular}{lcccc}
        \toprule
        \multirow{2}{*}{Iterations ($i$)} & \multicolumn{2}{c}{LV} & \multicolumn{2}{c}{Myo} \\
        \cmidrule(lr){2-3} \cmidrule(lr){4-5}
            & DSC $\uparrow$ & ASD $\downarrow$ & DSC $\uparrow$ & ASD $\downarrow$ \\\midrule
        
        $i=1$ & 64.85 & 16.26 & 51.20 & 13.81 \\
        $i=3$ & 74.05 & 10.68 & 57.83 & \textbf{9.48} \\
        $i=5$ & 73.94 & 11.60 & 58.66 & 10.42 \\
        $i=10$ & \textbf{76.93} & \textbf{8.77} & \textbf{59.43} & 11.68 \\
        \bottomrule
    \end{tabular}
    \end{adjustbox}
    \caption{Effect of the number of iterations of the proposed method during test-time adaptation on the UNet architecture for the ACDC $\mapsto$ LVQuant task. Evaluation metrics include the DSC (\%) and ASD (px), with the best-performing results highlighted in bold.}
    \label{tab:ablation_iteration}
\end{table}

%% file: Tables/running_time.tex

\begin{table}[tb]
    \centering
    \begin{adjustbox}{scale=0.8}
    \begin{tabular}{ccccc}
    \toprule
     Methods   &  UNet & MedNeXt & SwinUNETR \\\midrule
     TENT    & 0.18 \textcolor{green}{(-21.74\%)} & 0.19 \textcolor{green}{(-82.24\%)} & 0.18 \textcolor{green}{(-48.57\%)}\\
     CoTTA   & 1.76 \textcolor{red}{(+665.22\%)} & 3.47 \textcolor{red}{(+224.30\%)} & 1.99 \textcolor{red}{(+468.57\%)}\\
     TEA     & 0.25 \textcolor{red}{(+8.7\%)} & 4.15 \textcolor{red}{(+287.85\%)} & 0.63 \textcolor{red}{(+80.00\%)}\\
     Ours    & 0.23 & 1.07 & 0.35 \\
     \bottomrule
    \end{tabular}
    \end{adjustbox}
    \caption{\revise{Inference time per sample (in seconds) measured on a single NVIDIA RTX 2080 Ti GPU with 11 GB memory.}
    }
    \label{tab:running_time}
\end{table}